\documentclass[runningheads]{llncs}

\usepackage{eccv}

\usepackage{eccvabbrv}

\usepackage{graphicx}
\usepackage{booktabs}

\usepackage[accsupp]{axessibility}  % Improves PDF readability for those with disabilities.

\usepackage{hyperref}

\usepackage{orcidlink}

\usepackage{graphicx}	
\usepackage{amsmath}	
\usepackage{amssymb}	
\usepackage{booktabs}
\usepackage{epsfig}
\usepackage{caption}
\usepackage{float}
\usepackage{placeins}
\usepackage{color, colortbl}
\usepackage{stfloats}
\usepackage{enumitem}
\usepackage{tabularx}
\usepackage{xstring}
\usepackage{multirow}
\usepackage{xspace}
\usepackage{url}
\usepackage{subcaption}
\usepackage{xcolor}
\usepackage[hang,flushmargin]{footmisc}
\usepackage{seqsplit}

% ...

\newcommand{\SO}{\mathrm{SO}}
\newcommand{\SE}{\mathrm{SE}}

\newcommand{\Hcal}{\mathcal{H}}
\newcommand{\Ical}{\mathcal{I}}

\newcommand{\Ncal}{\mathcal{N}}

\newcommand{\btil}[1]{\mathbf{\tilde{#1}}}
\newcommand{\Rbb}{\mathbb{R}}

\newcommand{\R}[1]{{%
    \textbf{%
        \ifstrequal{#1}{1}{\textcolor{red}{R#1}}{%
        \ifstrequal{#1}{2}{\textcolor{blue}{R#1}}{%
        \ifstrequal{#1}{3}{\textcolor{magenta}{R#1}}{%
        \ifstrequal{#1}{4}{\textcolor{teal}{R#1}}{%
                           \textcolor{cyan}{R#1}%
        }}}}%
    }%
}}

\long\def\equivcvo{{\em EquivAlign }}
\long\def\equivcvoend{{\em EquivAlign}}

\newcommand{\zm}[1]{{#1}}
\newcommand{\comm}[1]{}

\newcommand{\lengthscale}{\ell}
\newcommand{\seThree}{$\mathrm{SE}(3)$ }
\newcommand{\soThree}{$\mathrm{SO}(3)$ }
\begin{document}
% \nolinenumbers
% ---------------------------------------------------------------
% TODO REVIEW: Replace with your title
\title{
%EquivCVO: 
%1.Correspondence-Free SE(3) Point Cloud Registration with Unsupervised Equivariance Learning  in Reproducible Kernel Hilbert Space
% \\
%2.Equivariance Learning in Reproducible Kernel Hilbert Space: A Unsupervised Correspondence-free Point Cloud Registration Framework
%\\
%3. Equivariant Kernel Learning for SE(3) Point Cloud Registration: A Unsupervised and Correspondence-free Approach \\
%4. Correspondence-Free SE(3) Point Cloud Registration  with Unsupervised Kernel Learning of Equivariance Features \\
 Correspondence-Free SE(3) Point Cloud Registration in RKHS via Unsupervised Equivariant Learning
}
% 

% \title{
% %EquivCVO: 
% Equivariance Registration in Reproducible Kernel Hilbert Space: An unsupervised correspondence-free Point Registration Framework}

% TODO REVIEW: If the paper title is too long for the running head, you can set
% an abbreviated paper title here. If not, comment out.
\titlerunning{EquivAlign: Corr.-Free Point Cloud Reg. via Unsupervised Equiv. Learning}

% TODO FINAL: Replace with your author list. 
% Include the authors' OCRID for the camera-ready version, if at all possible.
\author{Ray Zhang\inst{1}%orcidlink{0000-0001-9599-931X} 
\and
Zheming Zhou\inst{2}%\orcidlink{0000-0002-7549-1778} 
\and
Min Sun\inst{2} 
%\and
Omid Ghasemalizadeh\inst{2}
%\orcidlink{0000-0003-3142-8133} 
\and
Cheng-Hao Kuo\inst{2}
%\orcidlink{0000-0001-9464-9625} 
\and
Ryan M. Eustice\inst{1}
%\orcidlink{0000-0002-9989-4942} 
\and
Maani Ghaffari\inst{1}
%\orcidlink{0000-0002-4734-4295} 
\and
Arnie Sen\inst{2}
%\orcidlink{0009-0009-1786-1134}
}

% TODO FINAL: Replace with an abbreviated list of authors.
\authorrunning{R. Zhang et al.}
% First names are abbreviated in the running head.
% If there are more than two authors, 'et al.' is used.

% TODO FINAL: Replace with your institution list.
\institute{University of Michigan, Ann Arbor MI 48109, USA \\
\email{\{rzh,eustice,maanigj\}@umich.edu} \and
Amazon Lab126, Sunnyvale CA 94089, USA\\
\email{\{zhemiz,minnsun,ghasemal,chkuo,senarnie\}@amazon.com}}

\maketitle

\FloatBarrier
\vspace{-10pt}
\begin{abstract}

This paper introduces a robust unsupervised SE(3) point cloud registration method that operates without requiring point correspondences. The method frames point clouds as functions in a reproducing kernel Hilbert space (RKHS), leveraging SE(3)-equivariant features for direct feature space registration. A novel RKHS distance metric is proposed, offering reliable performance amidst noise, outliers, and asymmetrical data. An unsupervised training approach is introduced to effectively handle limited ground truth data, facilitating adaptation to real datasets. The proposed method outperforms classical and supervised methods in terms of registration accuracy on both synthetic (ModelNet40) and real-world (ETH3D) noisy, outlier-rich datasets. To our best knowledge,  this marks  the first instance of successful real  RGB-D odometry data registration using an equivariant method. The code is available at 
\url{https://sites.google.com/view/eccv24-equivalign}.

\end{abstract}

\section{Introduction}
\label{sec:intro}

\comm{
Point cloud registration estimates the relative transformation between two sets of 3D spatial observations~\cite{besl1992icp,chen1992surfacenormalicp,magnusson2007ndt3d, Zhang2020cvo, zhu2023e2pn}. It is usually formulated as a nonlinear optimization problem, whose inputs are the geometric and color point clouds captured by RGB-D cameras, stereo cameras, and LIDAR sensors. %,  containing rich intensity and geometric information. 
Various registration techniques have been successfully adopted in computer vision and robotics tasks such as visual odometry~\cite{Kerl2013DVOrepo} and 3D reconstruction~\cite{whelan2016elasticfusion}, but some open questions still remain, including nonlinear optimization on Riemannian manifolds, non-overlapped observations, perturbation from sensor noise and outliers, lack of high-quality initial pose guesses, etc. Different registration algorithms are determined by two tightly-coupled problems:  point representations and correspondences. Point representation is the actual form of raw point data used in the registration procedures, while the correspondences concern how the specific optimization residuals are constructed from all possible pairs of points.  They are correlated because the correspondence search is dependent on the chosen point representations.
}

Point cloud registration estimates the relative transformation between two sets of 3D spatial observations~\cite{besl1992icp,chen1992surfacenormalicp,magnusson2007ndt3d, Zhang2020cvo, zhu2023e2pn}. It is commonly formulated as a nonlinear optimization problem, with data inputs from varied sensors such as RGB-D cameras, stereo cameras, and LiDAR. This technique is vital in   computer vision and robotics, especially for applications like  visual odometry~\cite{Kerl2013DVOrepo} and 3D reconstruction~\cite{whelan2016elasticfusion}. Despite its wide use, point cloud registration encounters numerous challenges. These include complexities in nonlinear optimization on Riemannian manifolds, addressing non-overlapping observations, and mitigating the impact of sensor noise and outliers~\cite{yang2020teaser, qin2023geotransformer}.
These challenges stem from two tightly coupled components in traditional point cloud registration: point representations and correspondences. Point representation refers to the actual format of the point data in the process. Given a representation, point correspondences are related to the construction of the residuals from point pairings.
\begin{figure}[t]
    \centering
    \includegraphics[trim={1cm 3.5cm 1cm 2cm}, clip, width= 0.7\linewidth]{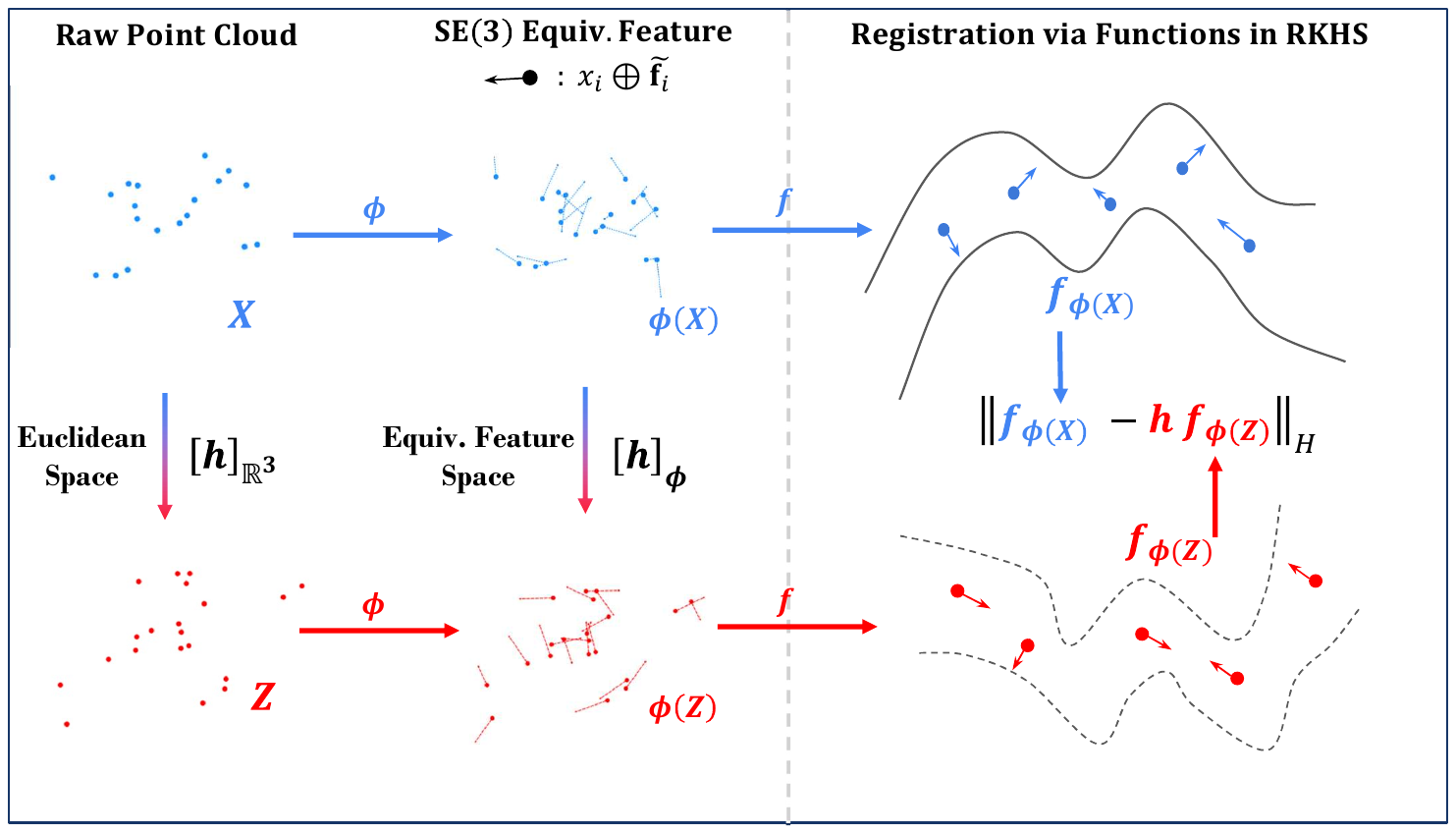}
    \vspace{-10pt}
\caption{%\textbf{Registration in RKHS with Equivariant Features:}  $\SE(3)$ registration of the two point clouds, $X$ and $Z$, in Reproducible Kernel Hilbert Space (RKHS). It directly takes  equivariant feature embeddings, $\phi(X)$ and $\phi(Z)$, instead of the raw 3D coordinates. The point clouds' feature embeddings  are further represented as  continuous functions $f_{\phi(X)}$ and $f_{\phi(Z)}$ in RKHS, which natively supports a distance metric between them, i.e. $||f_{\phi(X)} - f_{\phi(Z)}||^2_{\Hcal}$. To evaluate the distance metric, we will calculate the sum of kernels of point features, each represented as the direct sum $x\oplus \btil{f}$.  We can directly perform  pose optimization in the feature space because we can apply the transformation directly on the equivariant features.   
\textbf{Registration in RKHS with Unsupervised Learning of Equivariant Features:} The registration process takes  equivariant feature embeddings $\phi(X)$ and $\phi(Z)$ from point clouds $X=\{x_i\}\subset \mathbb{R}^3$ and $Z=\{z_j\}\subset \mathbb{R}^3$. The point cloud embeddings are represented as continuous functions $f_{\phi(X)}$ and $f_{\phi(Z)}$ in RKHS, allowing for the utilization of a distance metric, \mbox{$\| f_{\phi(X)} - h f_{\phi(Z)} \|^2_{\Hcal}$}, for direct estimation of the pose $h\in \SE(3)$  in the feature space. 
% Each point has the direct sum representation   $x_i \oplus \btil{f}_i$,  in the feature space, equivariant to rotations and translations.
\zm{In the feature space, each point is denoted as $x_i \oplus \btil{f}_i$ and represents the 3D coordinate, naturally exhibiting translation equivariance. This, combined with $\btil{f}_i$, the \soThree equivariant vectors, achieves \seThree equivariance.}
}
    \label{fig:teaser}
  \vspace{-20pt}
\end{figure}

\comm{
Classical registration methods' points are represented with hand-designed geometric primitives, like 3D point coordinates~\cite{besl1992icp, segal2009gicp},  planes~\cite{chen1992surfacenormalicp}, Gaussian mixtures~\cite{magnusson2007ndt3d, jian2011gmm}, and surfels~\cite{whelan2016elasticfusion,chen2019suma++}. These representations are in the form of low-dimensional vectors, and residuals can be directly calculated as Euclidean or Mahanobis distances.  However, they are constrained by the limited expressiveness of the real world's complex semantic structures, such as objects and textures. The limited expressiveness of this class of methods requires a sufficient number of correct data correspondence guesses as well as robust optimization procedures so as to minimize the effects of outliers~\cite{yang2020teaser}. 
}

Classical  methods represent points using hand-crafted geometric primitives such as 3D point coordinates~\cite{besl1992icp, segal2009gicp}, planes~\cite{chen1992surfacenormalicp}, Gaussian mixtures~\cite{magnusson2007ndt3d, jian2011gmm}, and surfels~\cite{whelan2016elasticfusion,chen2019suma++}. These representations, typically as low-dimensional vectors, allow residuals to be computed directly as Euclidean or Mahalanobis distances. However, they often struggle with handling noisy and outlier-rich observations because they rely on strict data correspondence. Correct data association is challenging~\cite{li2021generalized}, especially when the features are not discriminative enough. 
Robust optimization strategies are often needed to minimize these limitations~\cite{yang2020teaser}.

\comm{
For improved robustness and expressiveness, recent deep neural networks can learn point representations that capture geometric invariance~\cite{li2018pointcnn, dgcnn19,choy2019fully,qi2017pointnet}. It provides superior expressiveness for data association and robustness to noise and outlier perturbations in real benchmarks~\cite{yu2021cofinet,huang2021predator}. This class of methods aims at learning point-wise local and global features that stay invariant when poses act on them and hencewise enables semantic-aware data association. After the one-to-one correspondence is determined, RANSAC~\cite{fischler1981ransac} or weighted SVD are attached to regress the pose. %DCP~\cite{wang2019deep} and GeoTransformers~\cite{qin2023geotransformer} apply deep networks to 3D point cloud registrations.  
%However, to our best knowledge,   apart from associating classical CNN or MLP features, there are not much progress in associating dense equivariant features. In other words, if we extract the equivariant features, $f_1, f_2$ of two elements in the map, can we  directly compare their similarity with norms: $  ||f_1 - T f_2||$? 
However, existing generalization and invariant learning of registration algorithms require exploiting training data augmentation in the associations and in $\SE(3)$ poses. Real-world data is exposed to arbitrary transformations, and sampling in a six-dimensional space is expensive. 
}

In contrast, Continuous Visual Odometry (CVO)~\cite{MGhaffari-RSS-19, clark2021nonparametric, Zhang2020cvo } introduces a robust registration framework that represents each point cloud as a continuous function in RKHS. Its iterative registration process minimizes the distance of the two point cloud functions in RKHS and doesn't require strict pair-wise point correspondences. Although it demonstrates superior robustness compared to classical geometric registration methods, the iterative framework is constrained by limited expressiveness because the framework is not differentiable. 
% Specifically, in the iterative optimization procedure, the point cloud functions are transformed by the current estimate of the pose. Unlike the invariant feature-based method,  this step requires the point representations to respect the update of the pose.    

Advancements in deep neural networks bring differentiable point set registration by learning point representations that embody geometric invariance~\cite{qi2017pointnet,zeng173dmatch, li2018pointcnn, deng2018ppfnet, dgcnn19,choy2019fully} or equivariance.  \textit{Invariant-feature}-based approaches %offer enhanced expressiveness for data association and improved robustness against noise and outliers in real-world scenarios~\cite{yu2021cofinet,huang2021predator}. These methods 
focus on learning point-wise local and global features that remain invariant under pose transformations, leading to semantic-aware data association~\cite{yu2021cofinet,huang2021predator}. Once one-to-one correspondences are established, methods like RANSAC or weighted SVD are employed for pose regression~\cite{fischler1981ransac, besl1992icp}. However, challenges persist in the generalization of invariant learning. During training, excessive data augmentation is required for sampling transformations in  $\SE(3)$  and simulating the noise perturbations. Besides, their supervised nature rely on extensive ground truth labels. %Additionally, efficiently correlating with transformations in $\SE(3)$ space and managing the computational demands of sampling transformations in the $\SE(3)$ space remain significant hurdles%\ray{ These are required in only inference or both training and inference? Please also explicitly say what hurdles?}.
%\comm{
%Equivariant features~\cite{cohen2016group, cohen2016steerable,weiler20183dsteerable,thomas2018tfn}  provides an  alternative deep   representation for point clouds.   \textit{Equivariance} is a property for a map such that given a transformation in the input, the output changes in a predictable way determined by the input transformation: \begin{definition} A function $f: X \rightarrow Y$ is \textbf{equivariant} to a set of transformations $G$, if for any $g\in G$, $$g_Y  f(x) = f(g_X  x), \forall x \in X~.$$ \end{definition}  For example, applying a translation on a 2D image and then a convolution layer is identical to processing the original image with a convolution layer and then shifting the output feature map; hence, convolution layers are natively translation equivariant~\cite{cohen2016group}.  Recent developments of equivariant learning include $\SO(3)$~\cite{deng2021vectorneuron,zhu2022correspondence},  $\SE(3)$~\cite{ chen2021equivariant,zhu2023e2pn} and $\E(n)$~\cite{satorras2021n} equivariant networks that bring equivariance of more complex transformations to  3D point clouds. This property requires less data augmentation and labeling, thus leads to superior generalization~\cite{zhu2023e2pn} comparing to invariant learning-based  architectures. Most existing works are applied on applications on physics and chemistry, and simulated robotic datasets. Its performance on real point cloud registration datasets remain to be seen.  }

% On the other hand, 
\textit{Equivariant-feature}-based methods provide an alternative deep representation for point clouds~\cite{cohen2016group, cohen2016steerable,weiler20183dsteerable,thomas2018tfn}. Equivariance is a property for a map such that given a transformation in the input, the output changes in a predictable way determined by the input transformation: 
A function $f: X \rightarrow Y$ is equivariant to a set of transformations $G$, if for any $g\in G$, $g_Y  f(x) = f(g_X  x), \forall x \in X~.$
%For example, applying a translation on a 2D image and then a convolution layer is identical to processing the original image with a convolution layer and then shifting the output feature map; hence, convolution layers are natively translation-equivariant~\cite{cohen2016group}.
Recent strides in equivariant learning have expanded to include $\SO(3)$~\cite{deng2021vectorneuron,zhu2022correspondence}, $\SE(3)$~\cite{chen2021equivariant,zhu2023e2pn}, and $\text{E}(n)$~\cite{satorras2021n} equivariant networks. Compared to invariant feature-based methods, these networks relax the need for extensive data augmentation and thereby leading to improved generalization\cite{zhu2023e2pn}. % and less training time.%, and human annotation cost. %over invariant learning-based architectures.
While equivariant learning has shown promise within physics and chemistry, its effectiveness in real-world robotic tasks like point cloud registration is not well-established. For existing works, common practices include training a shape embedding to re-establish the one-to-one correspondence~\cite{zhu2022correspondence}, or pooling point-wise equivariant features into global equivariance features~\cite{chen2021equivariant, zhu2023e2pn}. These approaches undermine the complexities of the noisy and outlier-rich real data where the two input point clouds are not exactly the same, i.e., the equivariance does not fully hold.

In this work, we introduce an unsupervised feature space registration framework, \equivcvoend, %for \seThree point cloud registration, 
as depicted in Figure~\ref{fig_framework}. This framework focuses on learning point-wise representations that respect the intricate geometric structure in feature space. The proposed equivariant kernel learning interprets the neural feature embeddings of these point clouds as nonparametric functions within a   specified  RKHS.
This unique perspective allows for feature space registration without strict correspondences, further supporting the fully differentiable and unsupervised nature of our proposed method.
The contributions are outlined as follows:
\begin{enumerate}
\item An iterative and fully differentiable \seThree registration framework that facilitates correspondence-free feature space pose regression, enabling robustness to unseen noise and outliers.
%\item A lightweight equivariant encoder, supporting continuous rotational and translational equivariance.
\item A lightweight feature representation equivariant to 3D rotations and translations via a novel direct sum construction. This construction is modular and can easily benefit from future advances in equivariant encoders.
\item An unsupervised inner-outer loop training scheme for equivariant feature learning, incorporating a curriculum learning schedule, demonstrates enhanced accuracy compared to classical and supervised baselines and shows effectiveness in real-world applications. %where ground truth annotations are limited. %for iterative optimization in the equivariant feature space. The inner loop performs iterative pose optimization while the outer loop iteratively updates the equivariant encoder with curriculum learning.
%\item Validation of the proposed contributions' effectiveness on classical and learning-based point cloud registration baselines across synthetic and real-world datasets.

\end{enumerate}
\FloatBarrier
\vspace{-15pt}

%\zheming{my suggestion:}
%\begin{itemize}
%\item An iterative and fully differentiable \seThree registration framework that enables correspondence-free pose regression, demonstrating robustness to noise and outliers.

%\item A \seThree equivariant feature representation and learning approach within Reproducing Kernel Hilbert Space (RKHS). 

%\item An unsupervised inner-outer loop training strategy tailored for the \seThree registration task with real data is introduced within the framework, achieving superior accuracy on real-world datasets when compared to baseline methods.

%\end{itemize}

\section{Related Work}
\label{sec:related}

 \subsection{Classical Registration with ICP and GMM}

\zm{
The Iterative Closest Points (ICP) algorithm and its variants use hand-crafted geometric primitives like coordinates and surfaces as the point representation~\cite{besl1992icp, chen1992surfacenormalicp, mitra04_surface_icp, segal2009gicp}. They alternatively search correspondences with the closest geometric distances and then obtain pose estimates with the one-to-one data association.  Later works incorporate invariant features into the association for improved robustness, including color~\cite{servos2014mcicp, park2017colored}, intensity~\cite{parkison2019rkhsicp}, and semantic features~\cite{parkison2018semanticicp}. 

Gaussian Mixture Model (GMM) registration represents point clouds as probabilistic densities \cite{  jian2011gmm,campbell2015svr,chui2000mpm,eckart2018hgmr,evangelidis2017joint,horaud2010ecmpr}. Gaussian mixture models are fitted to the point cloud inputs, followed by soft data associations.  Normal Distribution Transform (NDT)  provides a particularly efficient way of modeling local geometric structures through  voxelization~\cite{biber2004probabilistic,magnusson2007ndt3d}.
}
% \begin{figure*}[t]
%   \centering 
% \includegraphics[width=\linewidth,trim={0cm 2.0cm 0.0cm 3.0cm},clip]{figs/framework_dark_camera_ready_v1.pdf}\vspace{-0.1cm}
% \caption{
% \textbf{\equivcvo Architecture:} An iterative, fully differentiable, and inner-outer loop structured unsupervised \seThree registration framework enables correspondence-free feature space pose regression. 
% During the training phase, the outer loop accumulates loss from the inner loop, which is dedicated to iterative pose adjustments aimed at refining the encoder. During the inference stage, raw point clouds are processed in a single pass by the encoder. Subsequently, the inner loop proceeds to iteratively optimize the pose within the feature space, continuing until convergence is reached.
% }
% \label{fig_framework}
%   \vspace{-10pt}
% \end{figure*}
\vspace{-10pt}
\subsection{Registration with Invariant Feature Matching}
Early works like FPFH~\cite{rusu2009fast} create histogram-based local invariant features that are used in global registration. 
% With the advances of deep learning, 
Deep invariant features provide a richer point representation that assists in feature space correspondence search. Encoders such as MLP~\cite{qi2017pointnet}, Graphic Neural Networks~\cite{dgcnn19}, and KPConv~\cite{thomas2019kpconv} are used for feature extraction that contribute to permutations invariance and local structures.  
In the correspondence step, direct supervision on inlier and outlier matches is usually required. This class of methods requires one-to-one pairwise matching, with either RANSAC or weighted SVD. To make the data association robust, complicated outlier rejection training mechanisms are adopted, assuming enough labeled training data. FCGF~\cite{FCGF2019} uses 
feature space metric learning with negative mining to filter the outliers. It samples 
both positive inliers and negative outliers so as to prevent the 
features being biased on the positive samples.   
Coarse-to-fine strategies in D3FeatNet~\cite{bai2020d3feat},  DCP~\cite{wang2019dcp},  Cofinet~\cite{yu2021cofinet}, PREDATOR~\cite{huang2021predator}, and GeoTransformer~\cite{qin2023geotransformer} enhance match precision by initially focusing on overlapping areas with superpixel or local patch matching, followed by finer point correspondences. Particularly, PREDATOR~\cite{huang2021predator} and GeoTransformer~\cite{qin2023geotransformer} leverage Graph Neural Networks (GNN) and cross-attention mechanisms for feature enhancement and adopt top-K neighbors for associations. These deep learning techniques, integrated with robust optimization methods like those in Teaser++~\cite{yang2020teaser}, represent a significant stride in achieving more accurate and reliable point cloud registration. However, their methods rely on costly labeling of ground truth and extensive data augmentation for generalization.

\vspace{-10pt}
\subsection{ Equivariant Learning and Applications in Registration}
%Equivariant learning respects the symmetry of group actions and neural network operations. %Classical 2D CNN is translational equivariant.

In the field of equivariant learning and registration, group convolution extends to various domains, beginning with Cohen's~\cite{cohen2016group} work on lifting convolution kernels to $\SO(2)$ rotations for image processing. This includes both discretizing rotations into finite groups like the dihedral group and continuous sampling with Monte Carlo~\cite{macdonald2022enabling}.
For 3D data, the icosahedron convolution theory~\cite{cohen2019gauge} and applications such as EPN~\cite{chen2021equivariant} and E2PN~\cite{zhu2023e2pn} leverage finite group discretization in point cloud analysis. These methods efficiently encode features across various angles in $\SO(3)$. Additionally, to learn translation-equivariant features, they incorporate traditional convolution layers.

Another approach involves continuous steerable feature maps in higher-order group representations, demanding significant computational resources for calculating coefficients~\cite{thomas2018tfn, cohen2018spherical,fuchs2020se3transformer}. VectorNeuron~\cite{deng2021vectorneuron} offers a more computationally efficient solution using only type-1 features. Existing equivariant methods with continuous group representations are mainly applied in physics and chemistry, whereas their performance in real robotics data requires further testing. Inspired by TFN~\cite{thomas2018tfn} and VectorNeuron~\cite{deng2021vectorneuron}, we construct a lightweight equivariant representation as a direct sum of point coordinates and \soThree steerable vectors to enable efficient translation and rotation equivariance.

\vspace{-10pt}
\subsection{Nonparametric Registration in RKHS}
% Apart from equivariant feature learning, another correspondence-free registration method represents points as continuous functions and registers directly in the function space. Continuous Visual Odometry (CVO)~\cite{ghaffari2019continuous}  proposes a new point cloud registration formulation that represents colored point
% clouds as continuous functions in RKHS and then aligns the
% two functions with gradient ascent.  The step size during the
% optimization is approximated with a fourth-order Taylor expansion. Kernel correlation~\cite{tsin2004correlation} is a special case of CVO that
% only performs geometric registration and optimizes the loss
% with a first-order approximation. AdaptiveCVO~\cite{lin2019adaptive} optimizes the kernel lengthscale of the original CVO.  SemanticCVO~\cite{Zhang2020cvo}  extends
% CVO to tightly couple hierarchical semantic information such
% as color or semantics with geometric information.  Comparing to CVO, the proposed work redesign the formulation to support differentiability, so as to learn the features specifically for the registration task. The features have to respect the iterative pose update during the inference stage, therefore equivariant features are necessary. In contrast, although SemanticCVO is able to utilize the deep features as well, it can only take the features from an existing network in an undifferentiable way. 

%Beyond equivariant feature learning, 
Continuous Visual Odometry (CVO)~\cite{ghaffari2019continuous} introduces a novel point cloud registration formulation by representing colored point clouds as continuous functions in an RKHS and aligns these functions using gradient ascent. The optimization step size is estimated through a fourth-order Taylor expansion. Kernel correlation~\cite{tsin2004correlation}, a specific instance of CVO, focuses solely on geometric registration and optimizes the loss using a first-order approximation. AdaptiveCVO~\cite{lin2019adaptive} 
%enhances the original CVO by 
optimizes the kernel length scale, while SemanticCVO~\cite{Zhang2020cvo} %expands CVO by 
integrates hierarchical semantic information, such as color and semantics, with geometric data.

\equivcvo sets itself apart from the above methods by reformulating the approach to ensure differentiability, thus enabling the learning of features meticulously designed for the registration task. Such features are required to support iterative pose updates at the inference stage, underscoring the necessity for equivariant features. In contrast, while SemanticCVO can also leverage deep features, it employs them in a non-differentiable fashion, depending on features derived from a pretrained network.

\vspace{-8pt}
\section{Problem Formulation 
}
% In preparation for delving into the details of our \equivcvo in the following section, it is beneficial to briefly review the notations and core principles of the problem.

Before delving into the proposed \equivcvo in the following section, we briefly review the notations and core principles of the problem.

% outlined in the CVO framework~\cite{clark2021nonparametric,Zhang2020cvo}.%\ray{ Can we move this to the beginning of Sec. 3?}

% \subsection{Problem Definition and Notations}
 
Consider two (finite) collections of points, $X=\{x_1, ..., x_N\}\subset \mathbb{R}^3$, $Z=\{z_1, ..., z_M\}\subset \mathbb{R}^3$, with $N,M$ not necessarily being equal. We aim to find an element \mbox{$h\in \mathrm{SE}(3)$}, 
which minimizes a distance metric between two point clouds $X$ and $hZ = \{hz_j\}$: 
\begin{equation}
\label{fomulation}
\hat{h}=\arg\min_{h\in\SE(3)} d(X, hZ).
\end{equation}
An $\SE(3)$ group element  $h=(R, t)$ with $R\in \SO(3), t\in \mathbb{R}^3$ acting on a point  $x\in \mathbb{R}^3$ is given by $hx=Rx + t$.

% \subsection{Registeration in Reproducible Kernel Hilbert Space}
%comment out
\comm{
We briefly review the registration algorithm in CVO~\cite{clark2021nonparametric,Zhang2020cvo}. We introduce how the two point clouds are represented and how the alignment is measured under such point representations.  The reader can refer to their technical reports~\cite{clark2021nonparametric,Zhang2020cvo} for further details.
}

\zm{
}

The point clouds,  $X$ and $Z$, are first represented as functions $f_X,f_Z:\mathbb{R}^3\to\Ical$  that live in some reproducing kernel Hilbert space (RKHS), denoted as  $(\Hcal,\langle\cdot,\cdot\rangle_{\mathcal{H}})$. The group action $\mathrm{SE}(3) \curvearrowright \mathbb{R}^3$ induces an action on the RKHS, $\mathrm{SE}(3) \curvearrowright \Hcal$,  denoted as $h.f(x) := f(h x)$. Inspired by this observation, we  set $h.f_Z := f_{h Z}$.  Furthermore,  each point might contain pose-invariant information in different dimensions, such as color or intensity,  described by a point in an inner product space, $(\mathcal{I},\langle\cdot,\cdot\rangle_{\mathcal{I}})$.  To represent pose-invariant information, we introduce two labeling functions, $l_X:X\to\Ical$ and $l_Z:Z\to\Ical$ for the two point clouds with $l_X(hx)=l_X(x), l_Z(hz)=l_Z(z)$, respectively. 
With the kernel formulation~\cite{bishop2006pattern}, the point cloud functions are
\begin{align}
\label{problem_functional}
% \begin{split}
	 f_X(\cdot) := \sum_{x_i\in X} \,  l_X(x_i) k%_{\ell}
 (\cdot,x_i), \quad 
	f_{h Z}(\cdot) := \sum_{z_j\in Z} \,  l_Z(z_j) k%_{\ell}
 (\cdot,h z_j),
% \end{split}
\end{align}
where the kernels are symmetric and positive definite functions. %parameterized by  \ray{kernel} hyperparameter $\ell$: 
$k: \Rbb^3 \times \Rbb^3\rightarrow \Rbb$.
To measure the alignment of the two point clouds given an isometry transformation $h\in\SE(3)$ that preserves norms, we can use the distance between the two point cloud  functions~\cite{clark2021nonparametric} 
\begin{align}
      d(f_X, f_{hZ}) = \| f_X - f_{h Z} \|^2_{\Hcal}
     = \langle f_X, f_X \rangle_{\Hcal} + \langle f_{Z}, f_{Z} \rangle_{\Hcal} - 2\langle f_X, f_{hZ} \rangle_{\Hcal} .
\end{align}

The distance is well-defined because RKHS is endowed with a valid inner product. With the \textit{reproducing property}~\cite{berlinet04rkhsbook}, each inner product becomes %can be further broken into
\begin{align}
    \nonumber \langle f_X,f_{h Z}\rangle_{\Hcal} =  \sum_{\substack{x_i\in X, z_j\in Z}} \,\langle  l_X(x_i), l_Z(z_j)\rangle k%_{\ell}
    (x_i,h z_j).  \label{eq:double_sum}
\end{align}
\FloatBarrier
\vspace{-20pt}
%\ray{$\rangle_{\Hcal}$: Do we want to use this subscript in Eq. 3 as well?}

\begin{figure*}[t]
  \centering 
\includegraphics[width=\linewidth,trim={0cm 2.0cm 0.0cm 3.0cm},clip]{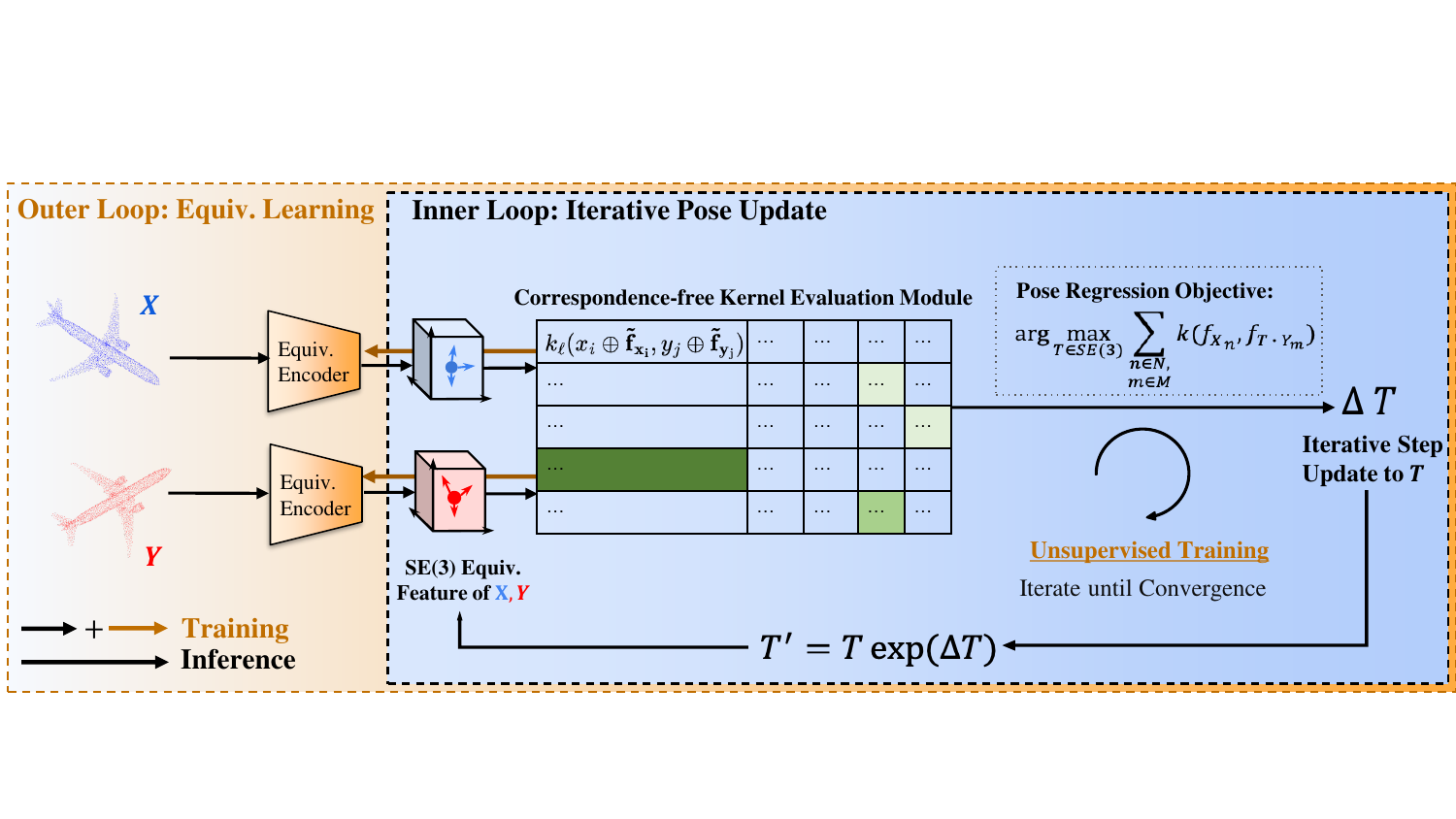}\vspace{-0.1cm}
\caption{
\textbf{\equivcvo Architecture:} An iterative, fully differentiable, and inner-outer loop structured unsupervised \seThree registration framework enables correspondence-free feature space pose regression. 
% In the training phase, the outer loop aggregates loss from the inner loop, which focuses on iterative pose updates to refine the encoder. During inference, raw point clouds are processed through the encoder a single time, following which the inner loop takes over, iteratively optimizing the pose directly within the feature space until convergence is reached.
During the training phase, the outer loop accumulates loss from the inner loop, which is dedicated to iterative pose adjustments aimed at refining the encoder. During the inference stage, raw point clouds are processed in a single pass by the encoder. Subsequently, the inner loop proceeds to iteratively optimize the pose that acts on the feature space, continuing until convergence is reached.
}
\label{fig_framework}
% \FloatBarrier
  \vspace{-20pt}
\end{figure*}
\section{\equivcvo Framework}
\label{sec:method}
\zm{
Figure~\ref{fig_framework} illustrates the \equivcvo framework. The process begins with the introduction of a lightweight \seThree equivariant feature representation as detailed in Section~\ref{subsec:equiv_repre}. Subsequently, we focus on optimizing the pose and kernel parameters within this feature space (Sections~\ref{subsec:kernellr} and~\ref{subsec:kernelchoice}). The training phase is distinctive due to the disparate stages and frequencies at which updates for the equivariant feature encoder, as well as for the kernel and pose, take place. To address this, we perform an unsupervised inner-outer loop learning strategy with curriculum learning, as discussed in Section~\ref{subsec:unsuptraining}. 
% During the inference phase, the encoder is fixed, allowing us to exclusively perform iterative inner loop pose updates. This process iteratively refines the pose between two point clouds until convergence is achieved.
}
\vspace{-10pt}
\subsection{Equivariant Point Representation}
\label{subsec:equiv_repre}
\begin{figure*}[t]
  \centering 
  %\resizebox{\textwidth}{!}{
  \includegraphics[width=0.99\textwidth%,trim={0.0cm 0.75cm 0.0cm 2cm},clip
  ]{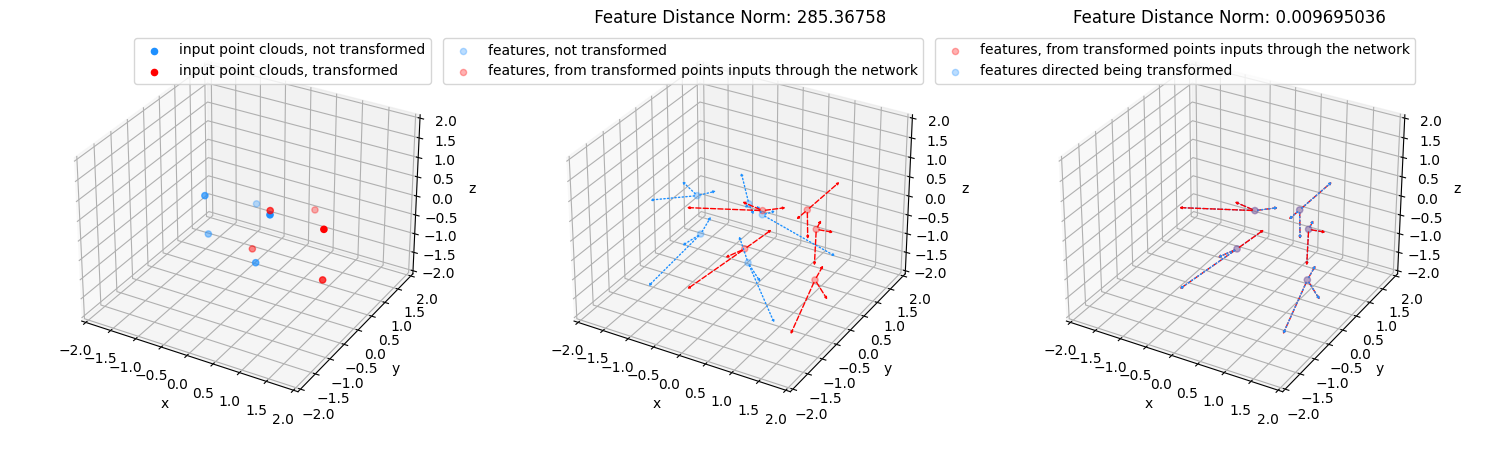}\vspace{-0.5cm}
  \caption{
  \zm{
    \textbf{SE(3)-Equivariant Representation of Point Feature:} (Left) Visualization of the two raw input point clouds in blue and red, being the 3D coordinate itself. (Middle) The direct sum representation of equivariant point features of the two point clouds at the initial relative pose, with each point appending its steerable vectors (for simplicity, three arrows per point are used in the illustration, representing three channels of each point's steerable features). (Right) Applied ground truth \seThree transformation to the feature space, resulting in an exact overlap of the two representations of the point set, affirming the precision of the equivariant representation. 
  }
  }
 \label{fig_equiv_example}
 %  \squeezeup\squeezeup

\FloatBarrier
\vspace{-10pt}
\end{figure*}

Unlike raw 3D coordinates, feature maps extracted from deep neural networks produce a more expressive representation of the point clouds. Instead of representing each point as an element in $\mathbb{R}^3$ as in CVO, we design equivariant features to represent them, $x\oplus \btil{f}$: a direct sum of $x$'s coordinate and multiple channels of $3$-dimensional steerable vectors $\btil{f}:=\phi(x)$, with  $\phi$ being the equivariant encoder with weights $\theta$.  The steerable features are a specific type-1 feature~\cite{thomas2018tfn} for rotations in  VectorNeuron~\cite{deng2021vectorneuron}. VectorNeuron proposes that $\SE(3)$-equivariance can be realized by centering the point cloud coordinates. However, in real applications,  the two input point clouds do not fully overlap. Thus, we cannot simply centralize them and process the rotation-only registration. Instead, we incorporate an additional type-0 feature—the 3D coordinate itself. Each channel of this representation can be visualized as a vector field defined on $\mathbb{R}^3$, as shown in Figure~\ref{fig_equiv_example}. This straightforward yet effective representation is modular, allowing for easy adaptation to future advancements in equivariant encoders.

The rotation $R$ and translation $t$ of the pose $h$ can be applied directly to the point-wise feature representations as follows:
\begin{align}
    R(x\oplus \btil{f}) = Rx\oplus R\btil{f}, \quad 
    t(x\oplus \btil{f}) = (t + x)\oplus \btil{f}.
\end{align}
The rotation's action on the features is on both the coordinates and the steerable vectors. The translation's action will only alter the coordinates but will not affect the vector field elements' directions.   

The linear multiplications by weights $W$ and the nonlinearity~\cite{deng2021vectorneuron} that acts solely on the steerable features can be defined as follows:
% We use the same  non-linearity as in VectorNeuron~\cite{deng2021vectorneuron}
\begin{align}
W(x\oplus \btil{f}) =  x\oplus  (\btil{f}W),\quad 
\sigma(x\oplus \btil{f}) =  x\oplus \sigma(\btil{f}).
\end{align}
The graph convolution over point $x$ is similar to DGCNN~\cite{dgcnn19} and VectorNeuron~\cite{deng2021vectorneuron}, but with the additional direct sum of 3D coordinate $x$ itself:%. The neighboring points include those whose steerable features are similar to the current point's in the feature space:
\begin{align}
W * (x\oplus \btil{f}) =  x\oplus \sigma(\btil{f}W +  \sum_{x_k\in \mathcal{N}(x)} (\btil{f}_k-\btil{f}) W_k)
\end{align} where $x_k\oplus \btil{f}_k$ are the neighbors' features and $W, W_k$ are the weights to learn. 
% \ray{$\oplus$ is element wise sum. direct sum is less clear to me.}
% The neighboring points include those whose steerable features are similar to the current point's in the feature space~\cite{dgcnn19}. 
\vspace{-10pt}
\subsection{Pose Optimization and Kernel Learning in the Feature Space}
\label{subsec:kernellr}
% To estimate the pose, we want to minimize the distance of the two functions in the RKHS:
To estimate the transformation, the objective is to minimize the distance between the two functions within the RKHS:
\begin{align}
    d(f_{\phi(X)}, f_{h\phi{(Z)}})=& \| f_{\phi(X)} \|^2 + \|f_{\phi(Z)}\|^2  - 2\langle f_{\phi(X)}, f_{h\phi(Z)} \rangle_{\Hcal}
\end{align}
where each function is represented as $f_{\phi(X)} = \sum l_X(x_i) k_{\ell}(x_i\oplus \btil{f}_i, \cdot)$. Let $\btil{f}_i:=\phi(x_i)$ and $\btil{g}_j:=\phi(  z_j)= \phi(z_j)$, then we have: 
\begin{align}
    \nonumber  &d(f_{\phi(X)}, f_{h\phi{(Z)}}) =    \sum_{i,j} \langle (l_X(x_i),l_X(x_j)\rangle k%_{\ell}
    (x_i\oplus \btil{f}_i,x_j\oplus \btil{f}_j) \\ \nonumber  &+ \sum_{i,j}  \langle l_Z(z_i),l_Z(z_j)\rangle k%_{\ell}
    (z_i\oplus \btil{g}_i, z_j\oplus \btil{g}_j) - 2\sum_{i,j} \langle  l_X(z_i),l_Z(z_j)\rangle k%_{\ell}
    (x_i\oplus \btil{f}_i, h(z_j\oplus \btil{g}_j))~.
\end{align} 
As the label function (color, intensity, etc.) remains invariant under the pose change, the following variables can be treated as constants: $c^X_{ij},c^Z_{ij},c_{ij}$. The final objective function becomes:
\begin{align}
    \nonumber d(f_{\phi(X)}, f_{h\phi{(Z)}}) = \sum_{i,j} c^X_{ij} k%_{\ell}
    (x_i\oplus \btil{f}_i,x_j\oplus \btil{f}_j) &+ \sum_{i,j} c^Z_{ij}k%_{\ell}
    (z_i\oplus \btil{g}_i, z_j\oplus \btil{g}_j)\\ - &2\sum_{i,j} c_{ij}k%_{\ell}
    (x_i\oplus \btil{f}_i, h(z_j\oplus \btil{g}_j)).
\end{align}

\vspace{-20pt}
\subsection{Kernel Choice}
\label{subsec:kernelchoice}
The RKHS, in which the point cloud functions reside, necessitates a well-defined Mercer kernel~\cite{berlinet04rkhsbook}. Characterized by a hyperparameter $\ell$, this kernel is a function of two variables within the equivariant feature space and has to be symmetric and positive-definite:  
% \begin{align}
$k_{\ell}: \phi \times \phi \rightarrow \Ical$.
% \end{align}
%\begin{align}
%    k_{\ell}(x,z) &= k_{\ell}(z,x)\\
%    \int k_{\ell}(x,z) g(x) g(z) du  &\geq 0  \,\,\,\, \forall g\in\Hcal
%\end{align}
 % and has to be symmetric and positive-definite. 
 
The selection of kernel is guided by two critical criteria. Firstly, it necessitates a minimal number of hyperparameters. In the classical CVO framework~\cite{Zhang2020cvo}, the kernel operates on three-dimensional inputs, where hyperparameters significantly influence the outcomes. Managing these hyperparameters becomes increasingly complex with higher-dimensional inputs, such as the direct sum of the 3D coordinate and the multi-channel steerable vectors. Secondly, the kernel's hyperparameters should be interpretable, enabling an understanding of its impact on the model's performance. %The current kernel choice links RBF kernel lengthscales to Euclidean point distances, whereas the $\tanh$ kernel only considers steerable vector directions. 
The kernel is defined as the product of the Radial Basis Function (RBF) kernel and the hyperbolic tangent kernel, as described  in~\cite{rasmussen2006gaussian}:
\begin{align}
    k_{\ell}(x_i\oplus \btil{f}_i, z_j\oplus \btil{g}_j) := \text{RBF}_{\ell}(x_i, z_j)\cdot \tanh{(1+\btil{f}_i\cdot \btil{g}_j)}. 
\end{align}
% because the product of two kernels is still a kernel. 
The RBF kernel is utilized for the coordinate part and the hyperbolic tangent kernel for the steerable feature maps. The RBF kernel includes a kernel parameter, the lengthscale $\lengthscale$, which is optimized during pose inference:
\begin{align}
    \text{RBF}_{\ell}(x_i , z_j )  = \exp(\dfrac{\|x_i  - z_j \|^2_3}{2{\lengthscale}^2}).
\end{align}
% while the hyperbolic tangent kernel does not. 
The RBF kernel is adopted to leverage the lengthscale parameter in promoting sparsity and minimizing the number of non-trivial terms in the loss calculation. A parameterized kernel is not selected for the steerable vectors  $\btil{f}$ to decrease the number of parameters requiring optimization during test time.

\subsection{Inference}
 During the inference stage, the goal is to minimize the distance between two functions with respect to the pose $h$ and the kernel parameter $\lengthscale$, while keeping the encoder weights $\theta$ fixed:
\begin{align}
\label{loss_full}
     \hat{h}, 
     \hat{\lengthscale}=\arg \min_{h, \lengthscale}d(f_{\phi(X)}, f_{h{\phi(Z)}}). %= \sum_{i,j} c^X_{ij} k_{\ell}(x_i\oplus \btil{f},x_j\oplus \btil{f}) \\ &+ \sum_{i,j} c^Z_{ij}k_{\ell}(z_i\oplus \btil{f}, z_j\oplus \btil{f}) - 2\sum_{i,j} c_{ij}k_{\ell}(x_i\oplus \btil{f}, z_j\oplus \btil{f})~.
\end{align} 
 % Note that for each iteration of the pose optimization, we don't need to resend the transformed point cloud through the encoder again. Instead, we just need to directly transform the equivariant features and re-evaluate the kernels in the loss. 
 It's important to note that for each iteration of pose optimization, there's no need to process the transformed point cloud through the encoder again. Instead, the approach involves directly transforming the equivariant features and re-evaluating the kernels during the loss calculation.

\subsection{Unsupervised Training of Equivariant Encoder}
\label{subsec:unsuptraining}

In practical scenarios such as visual odometry, ground truth transformation labels are often scarce. To adapt the encoder weights to new environments, unsupervised bi-level training~\cite{pineda2022theseus} is employed: 
\begin{align}
    \text{Inner Loop}: \arg\min_{h, \ell}d(f_{\phi(X)}, f_{h{\phi(Z)}}), 
    \text{Outer Loop}: \arg\min_{\theta}d(f_{\phi(X)}, f_{\hat{h} {\phi(Z)}}).
\end{align}
%\ray{$\min_{h, \ell}$ not the same symbol as in Eq. 12.}
 
% In training, we first send the two point clouds $X,Z$ through the equivariant encoder $\phi$ to obtain the point-wise equivariant features $\phi(X), \phi(Z)$. Then, in each iteration,  we minimize the loss with respect to the pose $h$ and kernel parameter $\ell$ to produce a step pose update. Based on the latest pose estimate $\hat{h}$, we keep the gradient in the computation graph and update the encoder parameters.  This training strategy doesn't require the ground truth pose label. 

During training, the two point clouds $X,Z$ are initially processed through the equivariant encoder $\phi$ to derive the point-wise equivariant features $\phi(X), \phi(Z)$. Subsequently, in each iteration, the loss is minimized with respect to the transformation $h$ and the kernel parameter $\ell$, facilitating a stepwise update of the transformation. Using the latest pose estimate $\hat{h}$, the gradient is retained in the computation graph, and the encoder parameters are updated. This training strategy does not require ground truth pose labels.

% We have some further considerations of the training procedure.  We use the curriculum training strategy to bootstrap the training when we start from random initial weights. We start from smaller angles at $1^{\circ}$ and gradually towards larger angles at $90^{\circ}$. Besides, the kernel parameter will occasionally change too fast and effectively become all zero. To prevent this from happening, we use a 100 times smaller learning rate for updating the kernel lengthscale $\lengthscale$.

Additional aspects of the training procedure are necessary to ensure satisfactory convergence properties. A curriculum training strategy is employed to initiate training from random initial weights, starting with smaller angles at $1^{\circ}$ and progressively advancing to larger angles up to $90^{\circ}$. Moreover, there is a tendency for the kernel parameter to change too rapidly, potentially causing its values effectively becoming zero. To mitigate this issue, a learning rate that is 100 times smaller is utilized specifically for updating the kernel lengthscale $\lengthscale$.

\section{Experiments}
% In this section, we present qualitative and quantitative experimental results on  the simulated ModelNet40 dataset~\cite{wu20153modelnet}  and real ETH3D RGB-D dataset~\cite{schops2019badslam}. We evaluate \equivcvoend's registration accuracy in rotations and translations, robustness, and generalization capability. For each dataset, we use the same set of hyperparameters. 

In this section, qualitative and quantitative experimental results are presented on both a simulated dataset, the ModelNet40 dataset~\cite{wu20153modelnet}, and a real-world RGB-D dataset, ETH3D~\cite{schops2019badslam}. The assessment focuses on \equivcvoend's registration accuracy in rotations and translations, along with its robustness to different perturbations. The implementation is based on Pytorch~\cite{paszke2017pytorch} and PyPose~\cite{wang2023pypose}. %Identical sets of hyperparameters are utilized for both datasets.

\textbf{Baselines:} Three types of baselines are chosen: a) Classical non-learning registration methods, including ICP~\cite{besl1992icp}, GICP~\cite{segal2009gicp} and the classical CVO~\cite{Zhang2020cvo}. For a fair comparison, CVO's label function $l_X(x_i)$ is set to $1$  to exclude extra information like color.  b)  Invariant feature-matching based methods, including RANSAC~\cite{fischler1981ransac} with FPFH features,  FGR~\cite{zhou2016fgr} with FPFH features, and GeoTransformer~\cite{qin2023geotransformer}.  GeoTransformer's official implementation is used, with the author-provided pretrained weights on ModelNet40 and our custom-trained weights on ETH3D. c) An equivariant feature method based on finite groups, E2PN~\cite{zhu2023e2pn}, trained under the same setup as \equivcvoend. 

\vspace{-10pt}
\subsection{Simulation Dataset: ModelNet40   Registration}
\label{subsec:modelnet}
\begin{figure}
     \centering
     \vspace{-20pt}
     %\subfloat[][Ground Truth Result]{\includegraphics[width=0.25\columnwidth,trim={0cm 0cm 0cm 0cm},clip]{figs/modelnet_gt_1.png}\label{fig_modelnet_gt}} 
     \subfloat[][$90^{\circ}$ Initial Rot.]{\includegraphics[width=0.25\columnwidth,trim={0cm 0cm 2cm 0cm},clip]{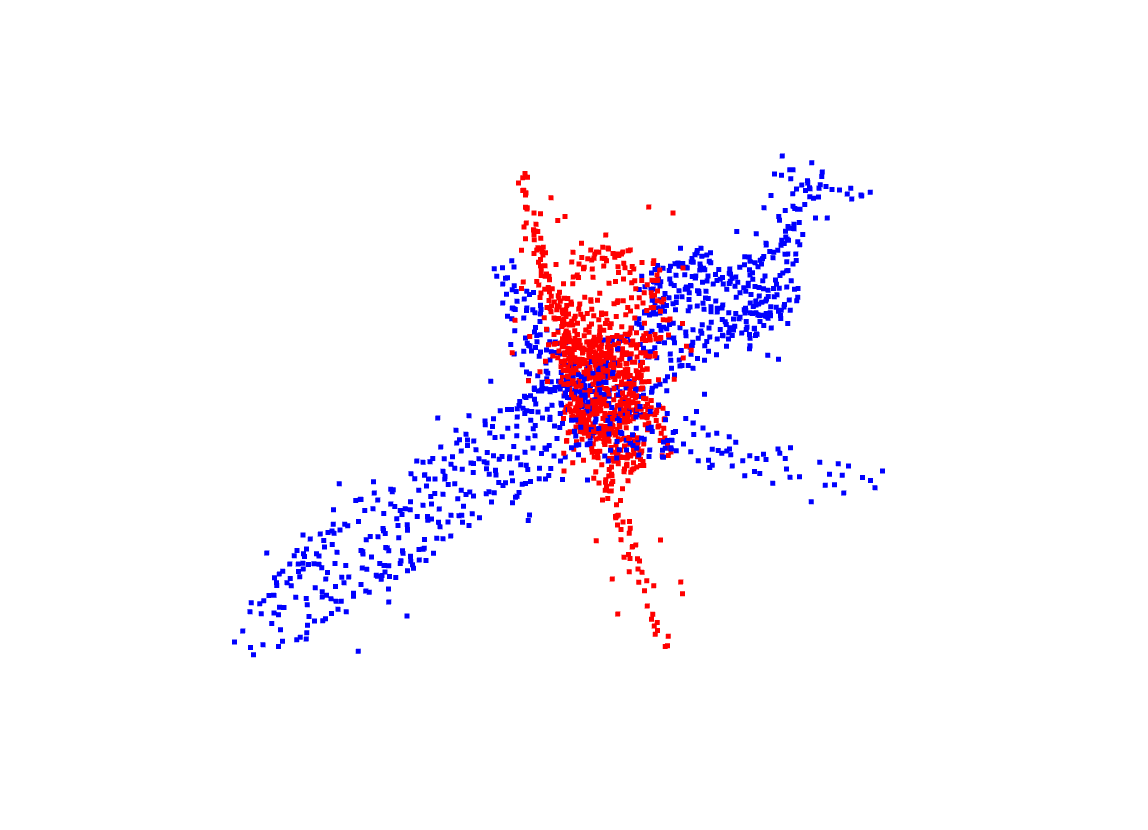}\label{fig_modelnet_init}} 
     \subfloat[][ICP]{\includegraphics[width=0.25\columnwidth,trim={0cm 0 0 0},clip]{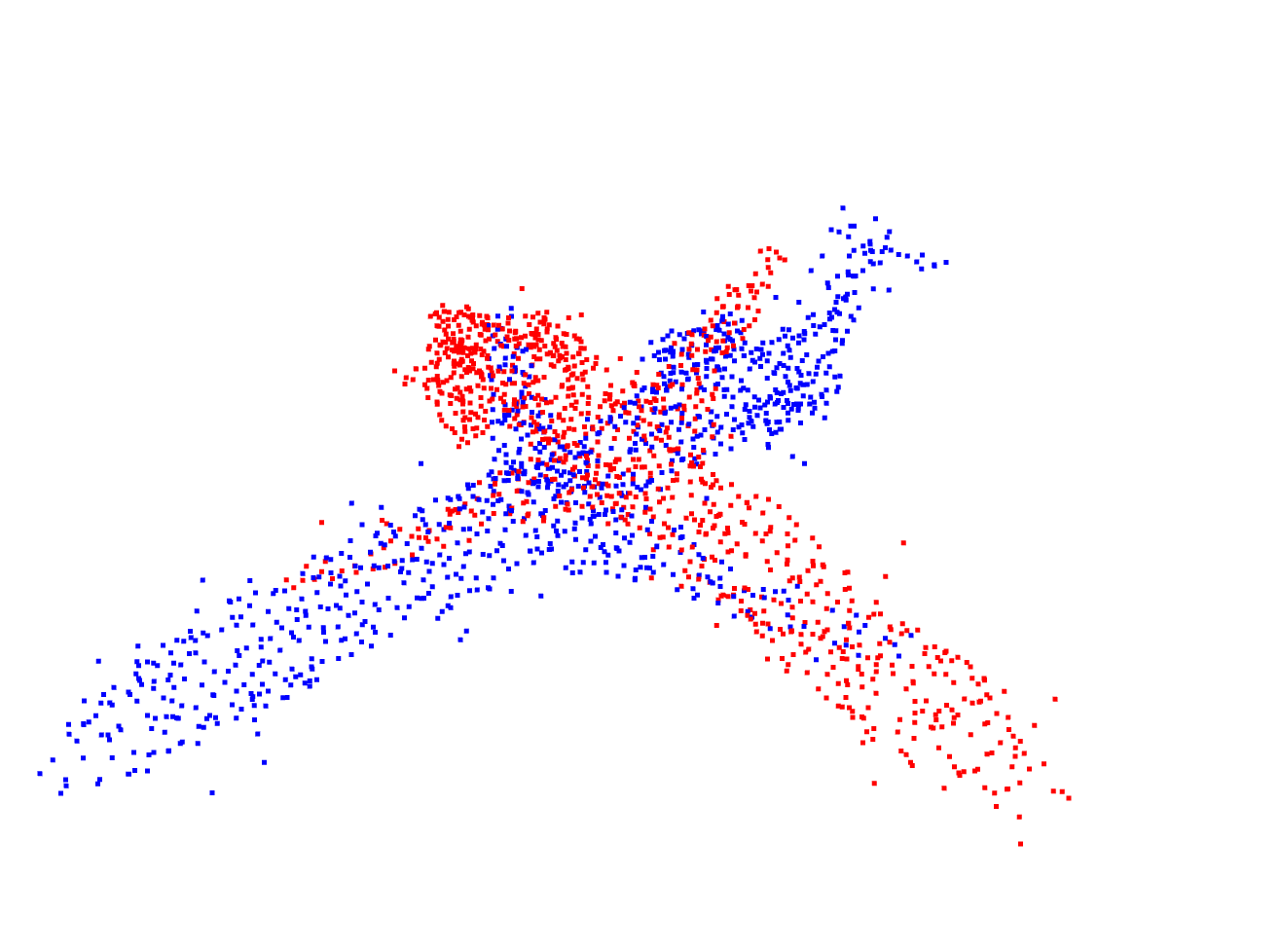}\label{fig:modelnet_icp}} 
     \subfloat[][GICP]{\includegraphics[width=0.25\columnwidth,trim={0cm 0 0 0},clip]{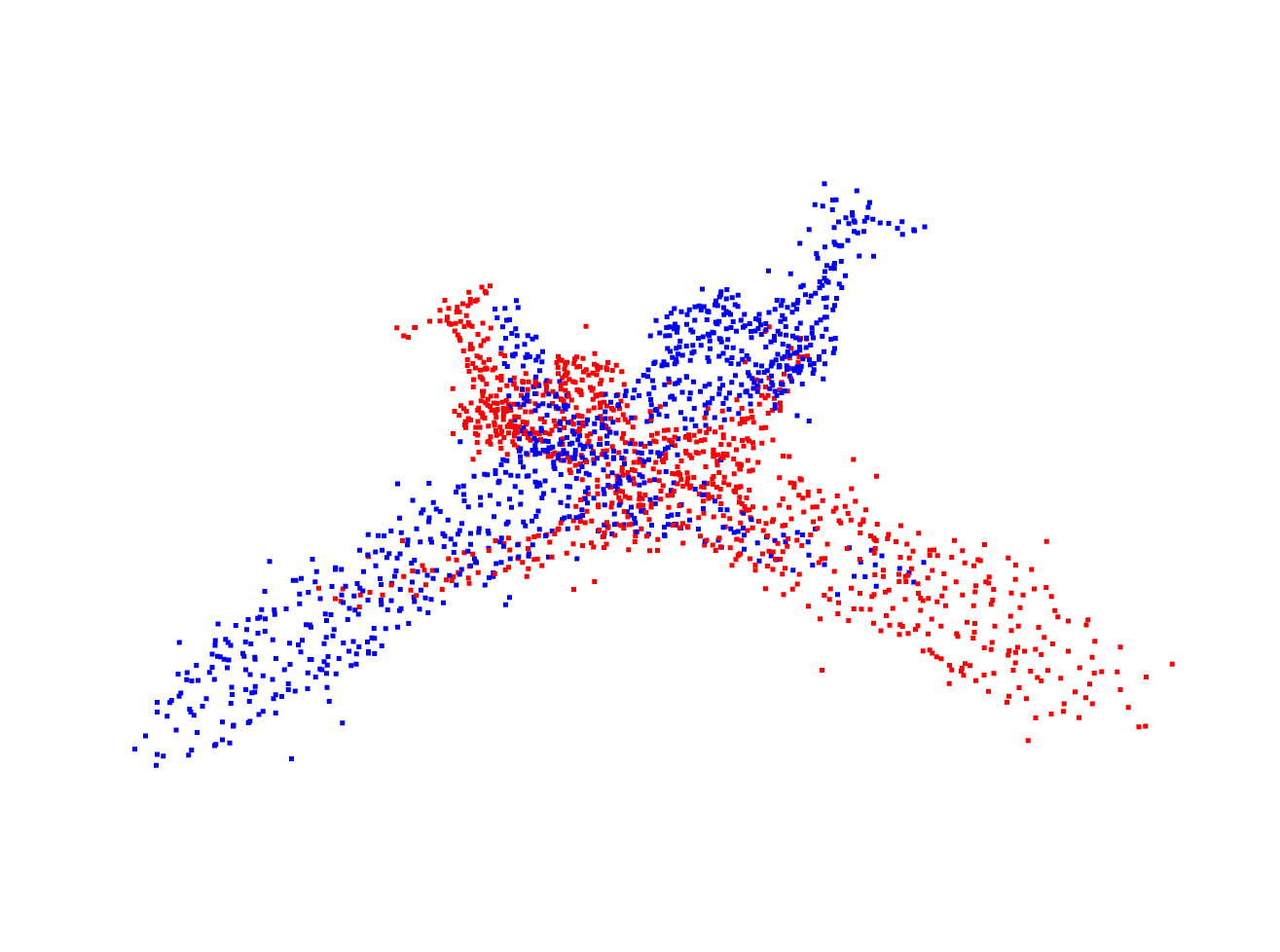}\label{fig:modelnet_gicp}} 
     \subfloat[][FPHF+RANSAC ]{\includegraphics[width=0.25\columnwidth,trim={0cm 0 5 0},clip]{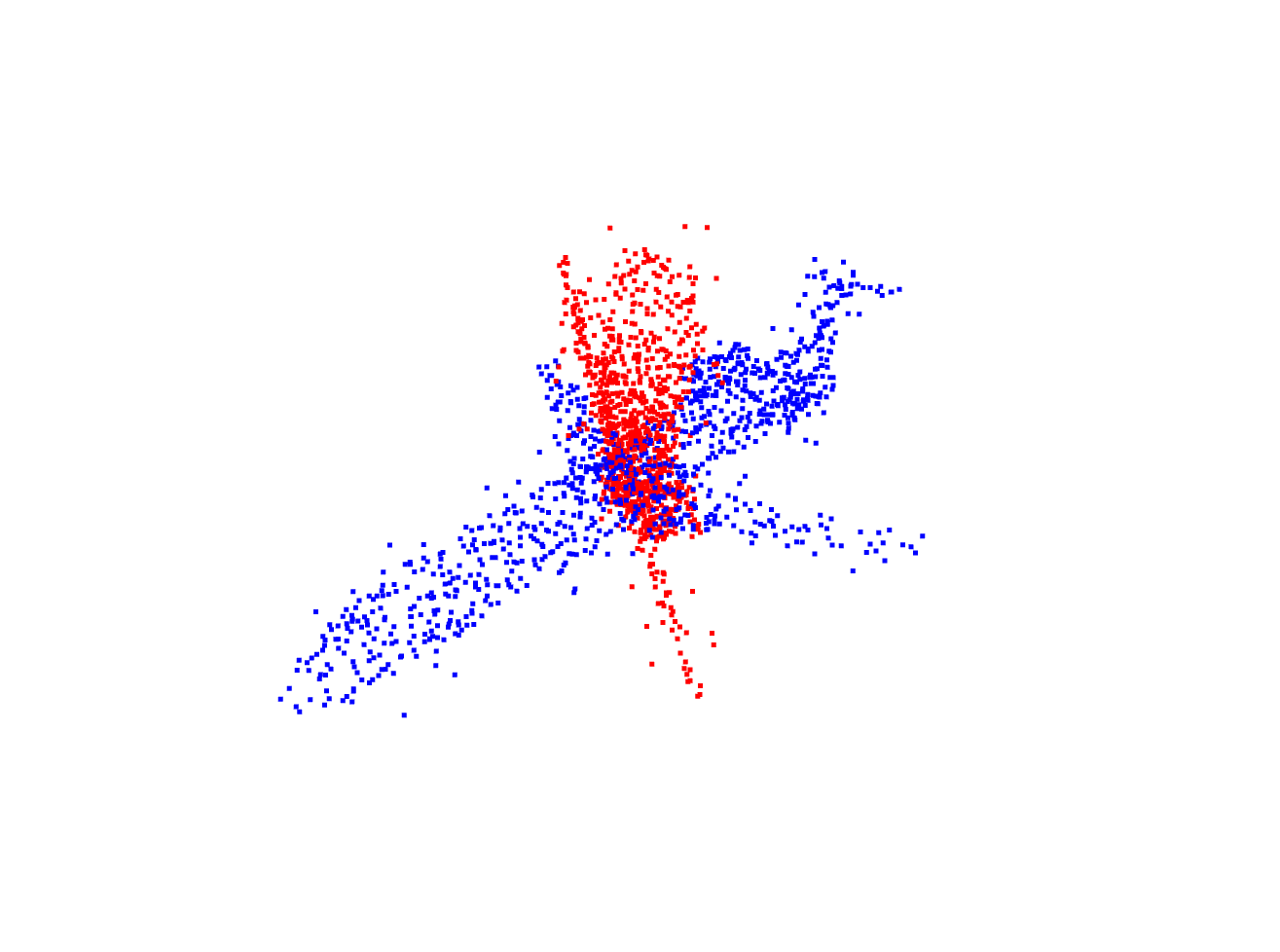}\label{fig:modelnet_fpfh_ransac}} \\
     \subfloat[][FPHF+FGR ]{\includegraphics[width=0.25\columnwidth,trim={0cm 0 0 0},clip]{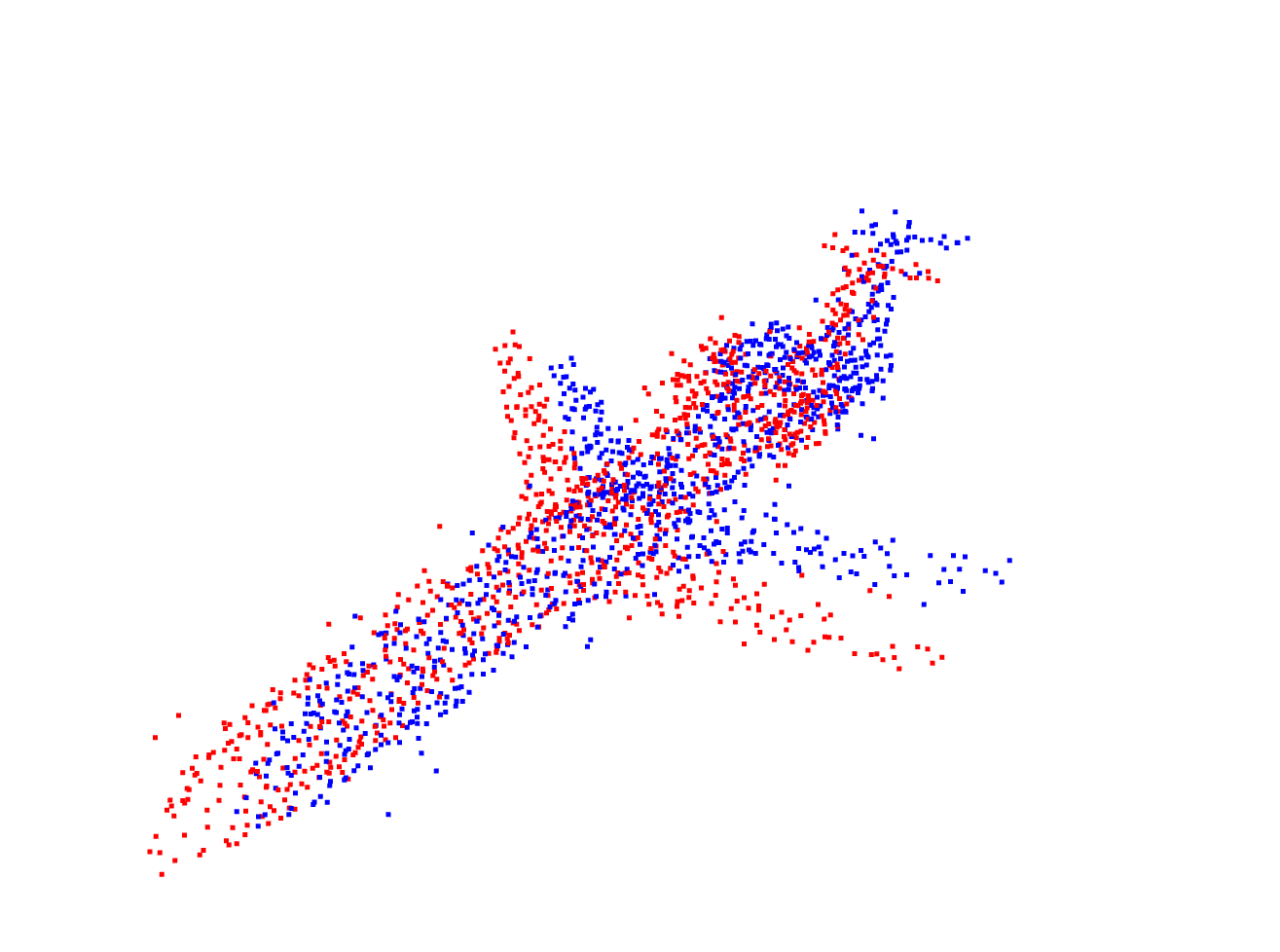}\label{fig:modelnet_fgr}} 
     \subfloat[][GeoTransformer]{\includegraphics[width=0.25\columnwidth,trim={0cm 0cm 0cm 0cm},clip]{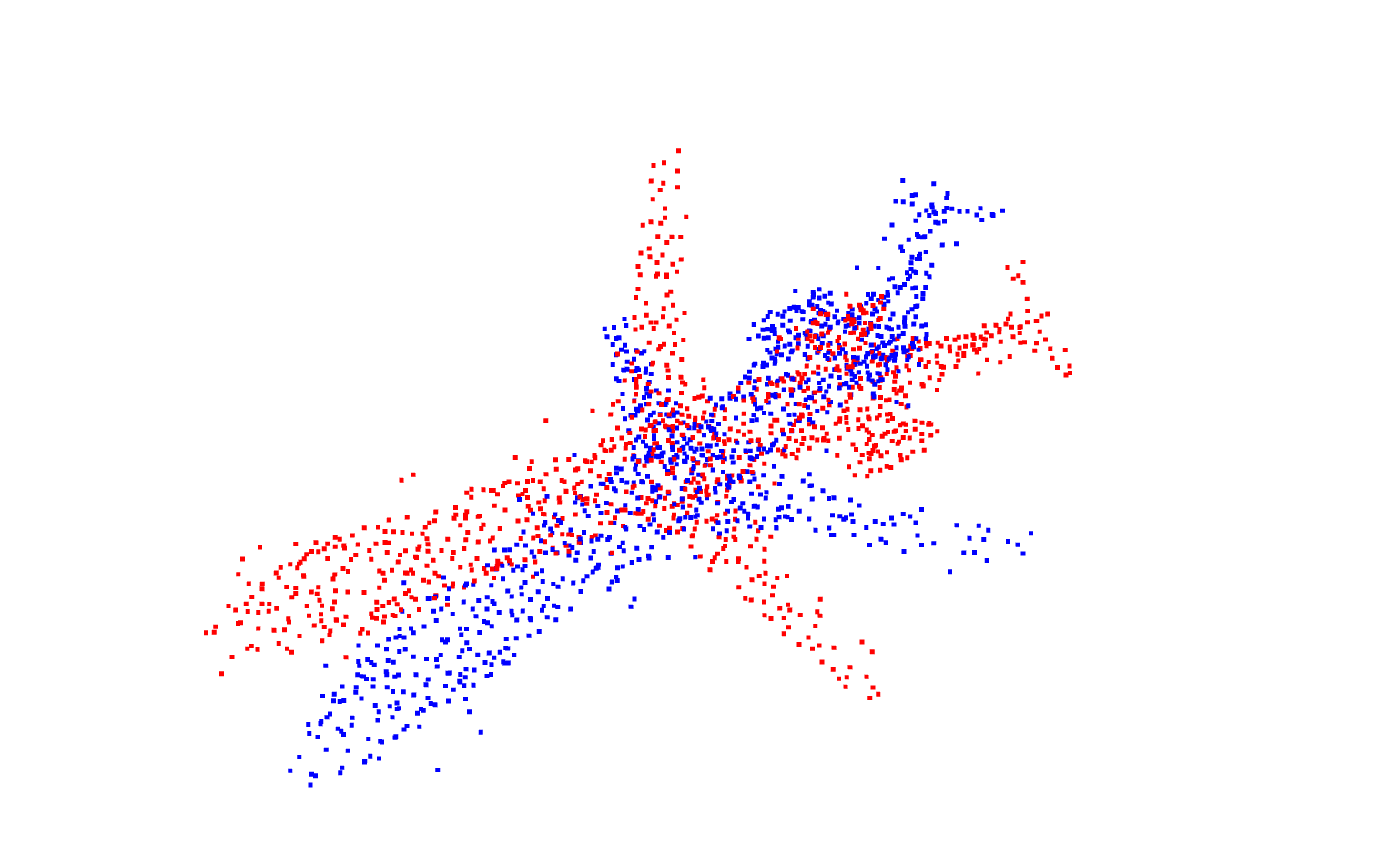}\label{fig:modelnet_geotransformer}} 
     \subfloat[][\zm{E2PN}]{\includegraphics[width=0.25\columnwidth,trim={0cm 0 0 0},clip]{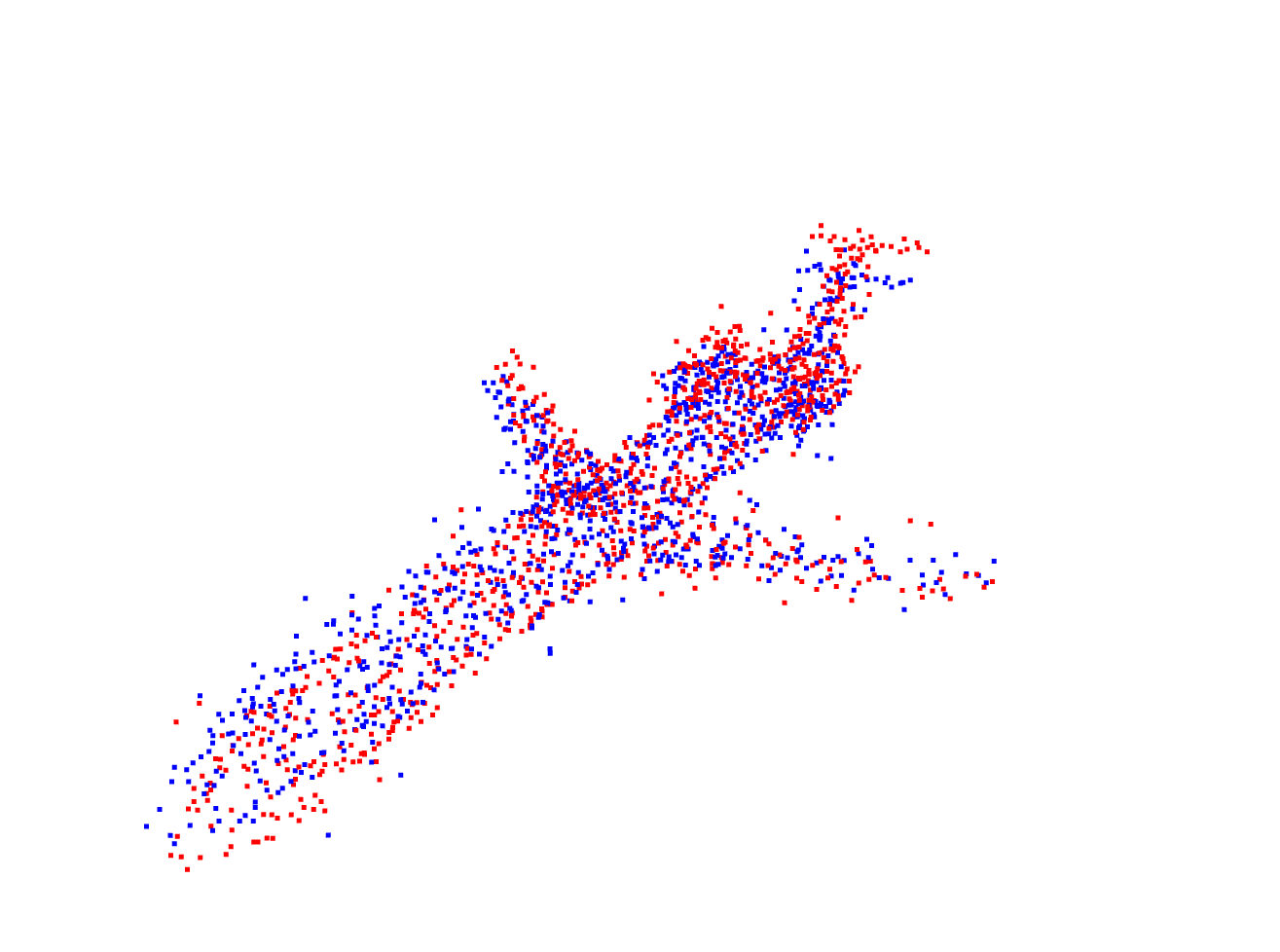}\label{fig:modelnet_e2pn}}
     \subfloat[][\equivcvoend]{\includegraphics[width=0.25\columnwidth,trim={8cm 1 1 1},clip]{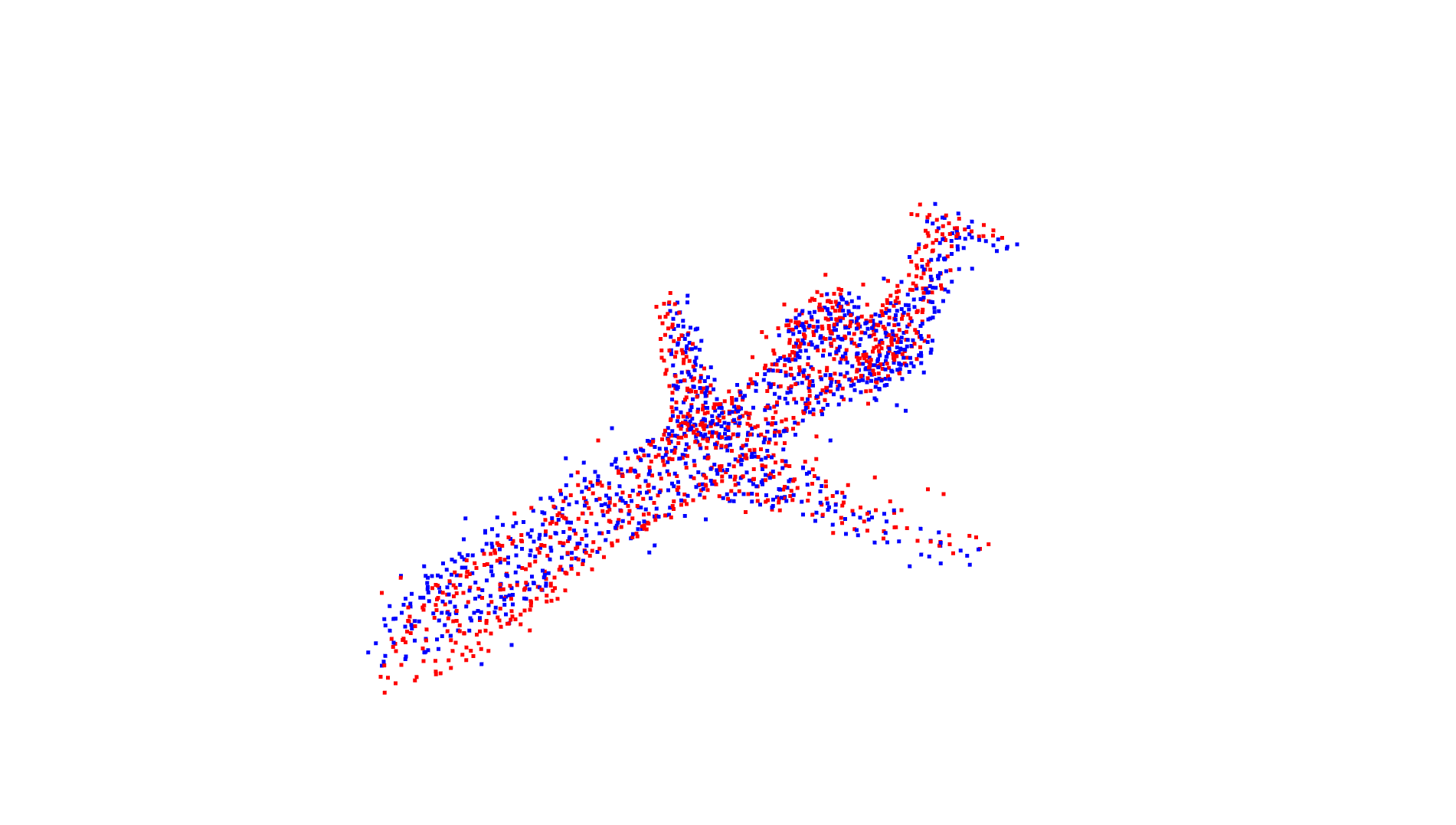}\label{fig:modelnet_equiv_cvo}}
     \caption{An airplane example of the point cloud registration at $90^{\circ}$ initial angle, with Gaussian noise $\Ncal(0,0.01)$ along the surface normal direction and $20\%$ uniformly distributed outliers. The equivariant registrations outperform the invariant and ICP-based methods. \equivcvo has a better yaw angle compared to E2PN. } 
     \label{exp_modelnet_qualitative}

\FloatBarrier
\vspace{-15pt}
\end{figure}

\textbf{Setup:}
In this experiment, we perform point cloud registration of all methods on the ModelNet40 dataset, which comprises shapes generated from 3D CAD models. To avoid the pose ambiguity of objects with symmetric rotational shapes, only the non-rotational symmetric categories   are
used in this experiment, with 60\%   training data, 20\% validation data, and
20\% test data. A point cloud is generated
by randomly subsampling 1,024 points on the surface, and
it is randomly rotated to form a pair. The initial rotation angle is set at $45^{\circ}$ and $90^{\circ}$ around random axes. The error metric is the mean of the matrix logarithm error between the resulting pose and the ground truth pose, i.e., $|| \log(h_{\text{result}} h_{\text{gt}}^{-1}) ||$. To assess the model's robustness under various noise perturbations, three types of noises are injected: a) Gaussian noise $\Ncal(0, 0.01)$ distributed along each point's surface normal. b) $20\%$ uniformly distributed outliers along each point's surface normal c) up to $20\%$ random cropping along a random axis. These noises are \textit{not} applied during training time for \equivcvo and E2PN. Note that only GeoTransformer includes these perturbations as data augmentations in its pretrained model.
 
%%%%%%%%%%%%%%%%%%%%%%%%%%%%%%%%%%%%%%%%%%%%%%%%%%%%%%%%%%%%%
% Table 1
\begin{table*}[t]
\centering
\resizebox{\textwidth}{!}{
\small
\begin{tabular}{l|c|c|c|c|c|c|c}
\toprule 

%top level
% \multicolumn{8}{c}{
% \zm{Results 1}
% } \\
% \midrule

\multirow{ 2}{*}{
%\rotatebox[origin=l]{90}{
Type 
}   & \multirow{ 2}{*}{
%\rotatebox[origin=l]{90}{
Method 
}  &  \multicolumn{3}{c}{Test Init Angle $<45^{\circ} $} & \multicolumn{3}{c}{Test Init Angle $<90^{\circ} $} \\ 
\cmidrule(lr){3-5} \cmidrule(lr){6-8}
&& $\sigma = 0, \gamma = 0$ & $\sigma = 0.01, \gamma = 0$ & $\sigma = 0.01, \gamma = 20\%$   & $\sigma = 0, \gamma = 0$ & $\sigma = 0.01, \gamma = 0$ & $\sigma = 0.01, \gamma = 20\%$ \\
\midrule
\multirow{ 2}{*}{
%\rotatebox[origin=l]{90}{
Non-Learning \bf
%}
} 
& ICP & 1.11  & 1.21 & 1.38&34.52 &
36.15 &  38.00\\
&GICP & 2.44  & 2.74 &2.53&49.88& 46.42 &48.40 \\
& Geometric-CVO & 5.67 & 5.93 &  6.28 &  23.90 & 26.92 & 30.84\\
\midrule
\multirow{ 2}{*}{
%\rotatebox[origin=l]{90}{
Invariant Features%}
} & FPFH + RANSAC &\textbf{0.02}&42.50&42.43&0.73
&85.63&85.60\\
& FPFH + FGR &0.07&2.27&12.62&\textbf{0.14}&11.69&43.88 \\
& GeoTransformer & 0.67&\textbf{0.71}&\textbf{0.92}&42.82&43.28&42.58 \\
\midrule
\multirow{ 2}{*}{
%\rotatebox[origin=l]{90}{
Equivariant Features
%}
}  & E2PN &3.86&46.84&70.88&3.78&48.19& 70.61 \\
&\equivcvo &0.61&
1.93 &
1.96&
1.14 %2.11
&
\textbf{4.08}&
\textbf{4.12}  \\

%%% Begin of the second table
\midrule
% \multicolumn{8}{c}{
% \zm{Results 2}
% } \\
\midrule

\multirow{ 2}{*}{
%\rotatebox[origin=l]{90}{
Type 
}   & \multirow{ 2}{*}{
%\rotatebox[origin=l]{90}{
Method 
}  &  \multicolumn{3}{c}{Test Init Angle $<45^{\circ} $} & \multicolumn{3}{c}{Test Init Angle $<90^{\circ} $} \\ 
\cmidrule(lr){3-5} \cmidrule(lr){6-8}
&  & crop $5\%$ & crop $10\%$ & crop $20\%$   & crop $5\%$ & crop $10\%$ & crop $20\%$ \\
\midrule
\multirow{ 2}{*}{
%\rotatebox[origin=l]{90}{
Non-Learning \bf
%}
} 
& ICP & 2.25 & 2.73 & 5.94 &37.00 &
39.23 &  45.04\\
&GICP & 3.33&3.34&5.68&49.28
&49.90&53.78 \\
& Geometric-CVO & 11.04 & 15.90 & 23.40 & 31.77 & 45.24 & 52.47 \\
\midrule
\multirow{ 2}{*}{
%\rotatebox[origin=l]{90}{
Invariant Features%}
} & FPFH + RANSAC &42.56&42.37&43.10&85.60
&85.51&85.17\\
& FPFH + FGR &37.93&44.73&57.15&78.66&82.22&90.94 \\
& GeoTransformer & \textbf{1.13}&\textbf{1.22}&\textbf{1.39}&41.41&42.04&45.45 \\
\midrule
\multirow{ 2}{*}{
%\rotatebox[origin=l]{90}{
Equivariant Features
%}
}  & E2PN & 76.90 &84.30 
& 92.41 & 76.06 & 83.42& 94.70 \\
&\equivcvo &9.88&
14.23 &
19.20&
\textbf{15.01} &
\textbf{28.87}&
\textbf{43.32}  \\
\bottomrule
\end{tabular}
}
\caption{
\zm{
\textbf{Rotation Error Analysis on the ModelNet40 Dataset:} (Top) Comparative performance of baselines under varying noise and outlier conditions. $\sigma$ is the variance of the Gaussian noise applied on the surface normal direction of each point. $\gamma$ is the ratio of points perturbed by uniformly distributed outliers. (Bottom) Baseline comparisons across different crop ratios. }
}
\label{exp_modelnet_benchmark}
\FloatBarrier
\vspace{-25pt}
\end{table*}

%%%%%%%%%%%%%%%%%%%%%%%%%%%%%%%%%%%%%%%%%%%%%%%%%%%%%%%%%%%%%
% Table 2
\comm{

\begin{table*}[t]
\label{exp_modelnet_crop}
\centering
\caption{Quantitative results}
\resizebox{0.75\textwidth}{!}{
\small
\begin{tabular}{l|c|c|c|c|c|c|c}
\toprule 
\multirow{ 2}{*}{
%\rotatebox[origin=l]{90}{
Type 
}   & \multirow{ 2}{*}{
%\rotatebox[origin=l]{90}{
Method 
}  &  \multicolumn{3}{c}{Test Init Angle $<45^{\circ} $} & \multicolumn{3}{c}{Test Init Angle $<90^{\circ} $} \\ 
\cmidrule(lr){3-5} \cmidrule(lr){6-8}
&& crop $5\%$ & crop $10\%$ & crop $20\%$   & crop $5\%$ & crop $10\%$ & crop $20\%$ \\
\midrule
\multirow{ 2}{*}{
%\rotatebox[origin=l]{90}{
Non-Learning \bf
%}
} 
& ICP & 1.21 & 1.84 & 4.61 &38.04 &
41.25 &  42.00\\
&GICP & 1.17&2.51&1.26&32.96
&37.14&41.27 \\
\midrule
\multirow{ 2}{*}{
%\rotatebox[origin=l]{90}{
Invariant Features%}
} & FPFH + RANSAC &42.83&42.75&43.10&85.89
&86.17&85.74\\
& FPFH + FGR &53.88&57.69&77.31&67.42&80.10&90.87 \\
& GeoTransformer &TODO&TODO&TODO&TODO&TODO&TODO \\
\midrule
\multirow{ 2}{*}{
%\rotatebox[origin=l]{90}{
Equivariant Features
%}
}  & E2PN &14.80&22.31
&18.76&24.19&18.05&21.80 \\
&\equivcvo &2.69&
7.07 &
14.48&
8.76 &
14.80&
30.10  \\
\bottomrule
\end{tabular}
}
\end{table*}

}
%%%%%%%% comment out original table
\textbf{Results:} The quantitative  results are  presented in Table~\ref{exp_modelnet_benchmark} and the qualitative results are shown in Figure~\ref{exp_modelnet_qualitative}. We denote the variance of the Gaussian noise as $\sigma$ and the ratio for the uniform outlier perturbation as $\gamma$.   

In noise-free conditions, both classical and proposed methods excel at smaller angles ($45^{\circ}$), with invariant feature-matching methods showing lower errors compared to equivariant-learning-based approaches. \equivcvo demonstrates performance on par with classical ICP methods and superior to E2PN. However, at initial angles of $90^{\circ}$, ICP and GICP show larger errors due to their reliance on accurate initial guesses for data association. In these scenarios, \equivcvo surpasses E2PN, but invariant feature-matching methods achieve the best results.

When encountering Gaussian noise, \equivcvo reaches a slightly better accuracy than the invariant feature matching methods at $45^{\circ}$ (except GeoTransformer) and is the best-performing method at $90^{\circ}$. GeoTransformer tops the benchmark at $45^{\circ}$.  Similar to the noise-less situation, non-learning methods' result degenerate at larger initial angles.

With $20\%$ uniformly distributed outliers, methods assuming Gaussian errors will degrade. Invariant feature matching is severely affected by this type of perturbation and fails to register at smaller or larger angles, unless extensive data augmentation is used during training. ICP-based methods can reach satisfactory results at small angles but not larger ones. 
% Equivariant learning-based methods are not heavily affected by this perturbation, while \equivcvo has an edge over the E2PN. 
\equivcvo remains largely unaffected by this perturbation, achieving the best results at $90^{\circ}$ by a significant margin and performing comparably to ICP-type methods at $45^{\circ}$.
This demonstrates how the expressiveness of equivariant features helps in the robustness of the registration process, even when only noise-free data is used in training. 

In tests involving random cropping of input data (with no cropping in training), as reported in Table~\ref{exp_modelnet_benchmark} (Bottom), all methods experience performance dips. Similar to the third case,  ICP-based methods are not substantially affected by the cropping at $45^{\circ}$ but are easily trapped in the local minima at larger angles. Classical invariant feature-based baselines cannot converge at either initial angle, but with sufficient data augmentation, GeoTransformer becomes the best-performing method. Both equivariant methods also experienced larger errors, not as severe as the classical invariant feature matching methods though. The proposed learning-based RKHS formulation natively annihilates the outlier disturbance because, at larger distances, the kernel will return trivial values. In contrast, as E2PN directly performs global pooling over all the points to obtain a single global feature, missing cropped components will reduce the quality of the global feature, especially when the crop is unseen in the training data. 

\vspace{-10pt}
\subsection{Real Dataset: ETH3D RGB-D Registration}
\label{sec:eth3d_exp}
%setskip
\setlength{\belowcaptionskip}{-10pt}

\begin{table}[t]
\centering
\begin{minipage}{.6\textwidth}
\resizebox{\textwidth}{!}{
\footnotesize
\begin{tabular}{l|c|cc|cc}
\toprule 
\multirow{ 2}{*}{
Type 
}   & \multirow{ 2}{*}{
Method 
}  &  \multicolumn{2}{c}{Rot. Error $(^{\circ})$} &  \multicolumn{2}{c}{Trans. Error $(m)$} \\ 
\cmidrule(lr){3-4} \cmidrule(lr){5-6}
&&   Mean & STD.   &  Mean & STD.  \\
\midrule
\multirow{ 2}{*}{
%\rotatebox[origin=l]{90}{
Non-Learning \bf
%}
} 
& ICP & 0.88 & 1.30 & 0.03 & 0.05 \\
& GICP & 0.69 & 3.54 & 0.02 & 0.11 \\
& Geometric-CVO & 0.71 & \textbf{0.94} & 0.02 & 0.02 \\
\midrule
\multirow{ 2}{*}{
%\rotatebox[origin=l]{90}{
Invariant Features%}
} & FPFH + RANSAC &8.75&2.95&0.17&0.40\\
& FPFH + FGR &3.60&12.61&0.08&0.17\\
& GeoTransformer &2.23&13.09&0.07&0.31\\
\midrule
\multirow{ 2}{*}{
%\rotatebox[origin=l]{90}{
Equivariant Features
%}
}  & E2PN &5.20& NA  
& NA & NA \\
&\equivcvo & \textbf{0.53} & 0.99 & \textbf{0.01} & \textbf{0.02}  \\
\bottomrule
\end{tabular}
}
\end{minipage}
%%%%%%%%%%%% image side by side  %%%%%%%%%%%%
\hfill
\begin{minipage}{.38\textwidth}
  \centering % Center the figure inside the minipage
  \resizebox{\textwidth}{!}{ % Adjust the resize box to fit within the minipage
  {\includegraphics[width=1.0\columnwidth,trim={1cm 0cm 3cm 3cm},clip]{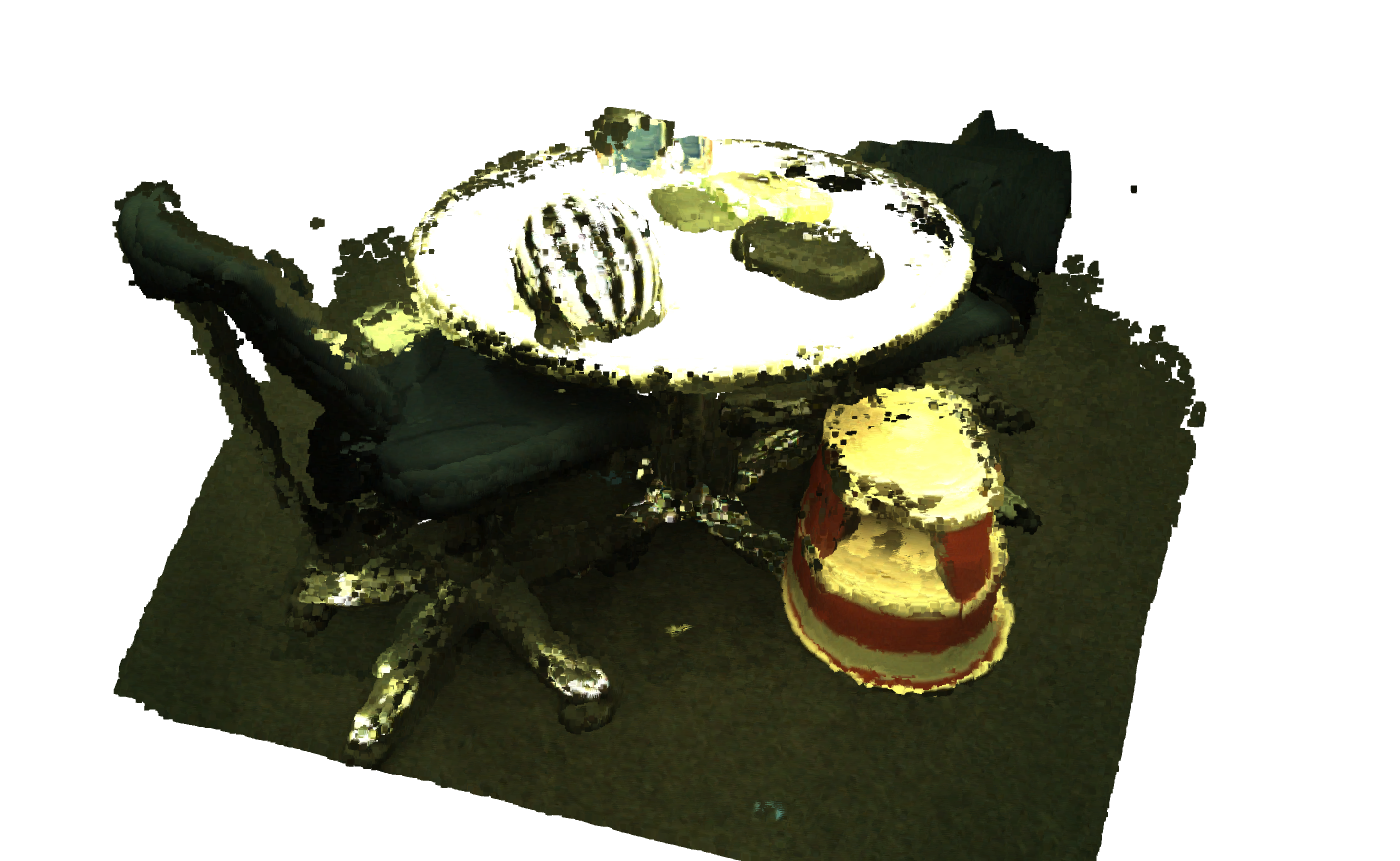}}
  } % Label for the Figure
  \label{exp_eth3d_qualitative}
\end{minipage}
% \begin{minipage}{0.4\textwidth}
%     \centering
%     \caption*{(a)}
%   \end{minipage}\hfill % Fill the space between the minipages evenly
%   \begin{minipage}{0.35\textwidth}
%     \centering
%     \caption*{(b)}
%   \end{minipage}
% \FloatBarrier
% \vspace{-10pt}
%%%%%%%%%%%% image side by side  %%%%%%%%%%%%

\caption{
% \textbf{Rotation and Translation Error Results on the ETH3D Dataset:} Among competing baselines, \equivcvo achieves the lowest rotation and translation errors. E2PN is SE(3) equivariant, but the official implementation does not include translation predictions for comparison. As a result, it is marked as "NA" in our table. 
\textbf{Quantitative and Qualitative Results on the ETH3D Dataset:} (Left) Among competing baselines, \equivcvo achieves the lowest rotation and translation errors. E2PN, while SE(3)-equivariant, lacks translation predictions in the official implementation and is marked as "NA" in our table. (Right)
Reconstruction using \equivcvo frame-to-frame transformations on the first 150 frames of \texttt{table\_3} sequence.
}
\label{exp_eth3d_benchmark}
\FloatBarrier
\vspace{-25pt}
\end{table}

%%%%%%%%%%%%%%%% Above are Table only  %%%%%%%%%%%%

% \begin{figure}[H]
%      \centering
%      \vspace{-10pt}
%        \subfloat[][\equivcvo reconstruction of \texttt{sfm\_table\_3}]{\includegraphics[width=0.5\columnwidth,trim={1cm 0cm 3cm 3cm},clip]{figs/table_3_reconstruct.png}}
%      \caption{\textbf{ Reconstruction Results on the ETH3D  Dataset}: Scene reconstruction results achieved by applying the frame-to-frame transformations estimated by \equivcvoend. Each reconstruction uses the first 150 frames of the corresponding sequence. 
%      } 
%      \label{exp_eth3d_qualitative}

% \end{figure}
% \FloatBarrier
% \vspace{-50pt}

%setskip 0
\setlength{\belowcaptionskip}{0pt}

\textbf{Setup:}
In this experiment, \equivcvo is benchmarked against other baselines using a real RGB-D dataset. We utilize the ETH3D dataset~\cite{schops2019badslam}, comprising real indoor and outdoor RGB-D images. In this setup, two point cloud pairs are sampled sequentially.  Unlike the simulated ModelNet40 dataset, a pair of point clouds will not fully overlap even without noise injections due to the viewpoint change. Additionally, the ground truth pose will contain rotation and translation but at smaller angles than the ModelNet40 experiment. A random rotation perturbation of $10^{\circ}$ is injected into each pair of testing data. 6 sequences are used for training: (\texttt{\seqsplit{cable\_3}}, \texttt{\seqsplit{ceiling\_1}},  \texttt{\seqsplit{repetitive}},  \texttt{\seqsplit{einstein\_2}},  \texttt{\seqsplit{sfm\_house\_loop}},  \texttt{\seqsplit{desk\_3}}),  2 sequences  for validation: (\texttt{\seqsplit{mannequin\_3}}, \texttt{\seqsplit{sfm\_garden}}), and 4 sequences for testing: (\texttt{\seqsplit{sfm\_lab\_room\_1}}, \texttt{\seqsplit{plant\_1}},  \texttt{\seqsplit{sfm\_bench}},  \texttt{\seqsplit{table\_3}}). Given the varying number of frames in each sequence, frame pairs are subsampled to ensure no more than 1000 pairs per sequence. This results in 5919 training instances, 2000 validation instances, and 2702 test instances. For all the methods, input point clouds are downsampled into 1024 points with the \texttt{\seqsplit{farthest\_point\_down\_sample}} method from Open3D~\cite{zhou2018open3d}.  For a fair comparison, color information is excluded in \equivcvo by setting the label function $l_X(x)=1$ in Eq.~\eqref{problem_functional} as the baselines similarly abstain from using color.

\textbf{Results:} The quantitative  results are  shown in Table~\ref{exp_eth3d_benchmark}. On the test sequences, \equivcvo demonstrates the best accuracy in both rotation and translation evaluations, with a $0.53^{\circ}$ rotation error,  $0.01m$ translation error, and lowest variations. The invariant feature-based baselines have significantly larger test errors. This is due to the challenge of generalizing to real-world noise, particularly for supervised learning methods. ICP-based methods have comparable translation errors, but their rotation error is $60\%$ and $25\%$ larger, respectively. This comparison indicates that \equivcvo produces fine-grained registration alone in real data and thus can be adopted in applications like frame-to-frame pose tracking. It does not have the necessity of using coarse-to-fine strategies with ICP, as adopted in recent invariant-learning-based works like  PREDATOR~\cite{huang2021predator}. 

Moreover, the other equivariant baseline, E2PN, is also not as accurate as \equivcvoend, though it is correspondence-free and has superior global registration ability. We argue that there are three potential reasons behind this: First, E2PN uses a finite group rotation representation on equivariance learning, resulting in a much faster running speed via feature permutation. However, the discretization comes at a cost; that is, it will have resolution challenges at fine-grained registration, especially compared to \equivcvoend's continuous rotation representation. Secondly, \equivcvo does not require training labels and thus is not tightly coupled to the training data distribution. In contrast, E2PN needs ground truth supervision, which means there would be overfitting challenges if the test set is a new scene. Thirdly, \equivcvo adopts the RKHS representation whose kernel can eliminate the influence of non-overlapped areas, while E2PN assumes complete symmetry of the input pair, which is often violated in real data. Recent works such as SE3-Transformer~\cite{fuchs2020se3transformer} and GeoTransformer~\cite{qin2023geotransformer} attempt to bring the attention mechanism to address this issue. But training the attention network will also need ground truth labels. 

% \FloatBarrier
% \vspace{-10pt}

% \FloatBarrier
\vspace{-10pt}

\subsection{Ablation Study}

\subsubsection{Kernel Choice}
The chosen kernel, although not our primary focus, prefers minimal hyperparameters and is interpretable, or any suitable kernel satisfying these. Even with a 3D kernel in the classical CVO~\cite{Zhang2020cvo}, the hyperparameters significantly impact the results. More complexity in controlling them could arise with higher dimensional inputs like the higher dimensional steerable vectors. The current kernel choice links RBF kernel lengthscales to Euclidean point distances, whereas the $\tanh$ kernel only considers steerable vector directions. To demonstrate that, we train and test the network with a single RBF kernel on ModelNet40 of 45$^{\circ}$ initial angles in Table~\ref{tab:ablation_tables} (a). 

% \FloatBarrier
% \vspace{-10pt}

% \begin{table}[h!]
% \centering
% \resizebox{0.3\columnwidth}{!}{
% \footnotesize
% \begin{tabular}{l|c}
% \toprule  
% % \textbf{ETH3D RGB-D Test}
% % }  
%  \textbf{Kernel Choice} &Rot. Error ($^{\circ}$) \\
%           &   Init Angle $<45^{\circ} $\\
% \hline
% RBF$\times$Tanh kernel & 0.29 \\
% RBF kernel only & 81.97 \\
% \bottomrule
% \end{tabular}
% }
% \caption{This table compares the RBF$\times\tanh$ kernel with RBF kernel alone for the equivariant features. After both are trained on the ModelNet40 dataset of initial angles under $45^{\circ}$, the RBF kernel alone cannot effectively complete the registration while the proposed RBF$\times\tanh$ kernel can.}
% \label{tab_kernel_choice}
% \end{table}

\FloatBarrier
\vspace{-10pt}
\subsubsection{Initial Kernel Parameter}

% \comm{
% \begin{table}[t]
% \centering
% \resizebox{0.35\textwidth}{!}{
% \footnotesize
% \begin{tabular}{l|c|c|c|c}
% \toprule 
  
% Init lengthscale ($\lengthscale$)  
% &  0.25 & 0.5   &  0.75 & 1.0  \\
% \midrule
% Init Angle: $45^{\circ}$  
%   & 48.2 & 1.12
%  & 0.93 &
% 0.29
% \\
% Init Angle: $90^{\circ}$ & 68.01 &
% 8.77
% & 5.95 &
% 1.07\\
% \bottomrule
% \end{tabular}
% }
% \caption{

% \textbf{Ablation Study on Kernel Lengthscale:} This study examines the effects of four distinct kernel lengthscales on two initial angles. The findings suggest a positive correlation between the problem scale and the kernel lengthscale, indicating that larger initial errors necessitate a correspondingly larger lengthscale.
% }

% \label{exp_ablation_with_lengthscale}
% \end{table}
% }

%% Kernel parameter plot
% \begin{figure*}[t]
%   \centering 
%   %\resizebox{\textwidth}{!}{
%   \includegraphics[width=0.4\textwidth%,trim={0.0cm 0.75cm 0.0cm 2cm},clip
%   ]{figs/ablation_rotation_ell.png}
%   \caption{
%  This study examines the effects of four distinct kernel lengthscales on two initial angles. The findings suggest a positive correlation between the problem scale and the kernel lengthscale, indicating that larger initial errors necessitate a correspondingly larger lengthscale.
%   }
%  \label{fig_ablation_ell}
% \end{figure*}

\equivcvo has a hyperparameter, the kernel lengthscale $\lengthscale$, which controls the coarse-grain and fine-grain resolution of the loss~\cite{clark2021nonparametric, Zhang2020cvo}. It is optimized during pose regression but still requires an initial value. In this ablation study shown in Table~\ref{tab:ablation_tables} (b), we test how the initial lengthscale will affect the registration accuracy on the ModelNet40 dataset. The outcome corroborates insights from the CVO works: Registering at larger angles necessitates a greater initial lengthscale for a comprehensive global perspective.
\FloatBarrier
\vspace{-10pt}
\subsubsection{Curriculum Learning vs. Direct Training}
As an unsupervised approach, one significant challenge we addressed was the bootstrap of the randomly-initialized network weights. To investigate this issue, we conducted an ablation study between curriculum-based training and direct training at maximal angular perturbations, as presented in Table~\ref{tab:ablation_tables} (c). Our findings show that a gradual curriculum with incremental steps significantly enhances model accuracy and reduces uncertainty, highlighting the effectiveness of curriculum learning for optimizing network performance from randomly initialized weights.

\begin{table}[t!]
\centering % Center the whole table

% Begin of the first minipage for Table1
\begin{minipage}{.28\textwidth}
  \centering % Center the table inside the minipage
  \resizebox{\textwidth}{!}{ % Adjust the resize box to fit within the minipage
  \footnotesize
  \begin{tabular}{l|c}
  \toprule
   \textbf{Kernel Choice} & Rot. Error ($^{\circ}$) \\
           & Init Angle $<45^{\circ} $\\
  \hline
  RBF$\times$Tanh kernel & 0.29 \\
  RBF kernel only & 81.97 \\
  \bottomrule
  \end{tabular}
  }
  % \textbf{(a)} % Label for Table1
\end{minipage}% End of the first minipage
\hfill % Space between the minipages
%for image
\begin{minipage}{.32\textwidth}
  \centering % Center the figure inside the minipage
  \resizebox{\textwidth}{!}{ % Adjust the resize box to fit within the minipage
  \includegraphics[width=\textwidth]{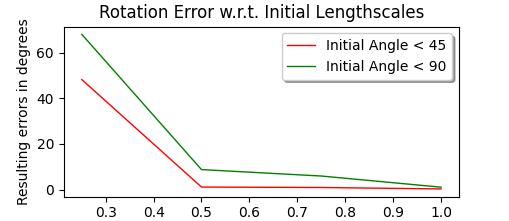}
  }
  % \textbf{(b)} % Label for the Figure
\end{minipage}% End of the third minipage
\hfill % Space between the minipages
% Begin of the second minipage for Table2
\begin{minipage}{.38\textwidth}
  \centering % Center the table inside the minipage
  \resizebox{\textwidth}{!}{ % Adjust the resize box to fit within the minipage
  \footnotesize
  \begin{tabular}{l|cc|cc}
  \toprule
    \multirow{ 2}{*}{Curriculum}  &  \multicolumn{2}{c}{ $[45^{\circ}]$} &   \multicolumn{2}{c}{ $[1^{\circ}, 10^{\circ}, 20^{\circ}, 30^{\circ}, 45^{\circ}]$ }\\  
  \cmidrule(lr){2-3} \cmidrule(lr){4-5}
   &  Mean & STD.   &  Mean & STD.  \\  \midrule 
  Init Angle: $45^{\circ}$ &  2.73  &  9.1 & 0.29 &
  0.469\\
  \bottomrule
  \end{tabular}
  }
  % \textbf{(c)} % Label for Table2
\end{minipage}% End of the second minipage

  \begin{minipage}{0.28\textwidth}
    \centering
    \caption*{(a)}
  \end{minipage}\hfill % Fill the space between the minipages evenly
  \begin{minipage}{0.32\textwidth}
    \centering
    \caption*{(b)}
  \end{minipage}\hfill % Fill the space between the minipages evenly
  \begin{minipage}{0.38\textwidth}
    \centering
    \caption*{(c)}
  \end{minipage}
% \FloatBarrier
% \vspace{-10pt}
% Single caption for both tables
\caption{\textbf{(a) Kernel Comparison: }
The table contrasts the RBF$\times\tanh$ kernel with the RBF kernel alone on equivariant features. Training on the ModelNet40 dataset for initial angles up to $45^{\circ}$, only the RBF$\times\tanh$ kernel successfully completes registration, unlike the RBF kernel alone. \textbf{(b) Lengthscale Study:} An analysis of four kernel lengthscales across two initial angles reveals a direct relationship between problem scale and lengthscale, suggesting larger initial errors require larger lengthscales. \textbf{(c) Necessity of Curriculum Learning:} Starting with small, incremental angles on the ModelNet40 dataset yields lower error means and STDs than starting from larger angles, underscoring curriculum learning's efficacy in training.}
\label{tab:ablation_tables}
% \FloatBarrier
 \vspace{-25pt}
\end{table}

\FloatBarrier
\vspace{-10pt}
\section{Limitations and Conclusion}
\label{sec:conclusion}
% The primary limitations of our method include lengthy training times and reduced performance in situations with large initial angles and limited overlap. First, our unsupervised training approach relies on a curriculum learning strategy to initiate the training process. This strategy, which gradually progresses from smaller to larger rotations, results in extended training durations. Specifically, training on the ModelNet dataset with 8 NVIDIA Tesla V100 GPUs takes about a week for $90^{\circ}$ registrations. Second, in the absence of direct supervision for overlap, our method exhibits decreased performance in low-overlap scenarios. For instance, as indicated in Table~\ref{exp_modelnet_benchmark}, when the overlap region is as low as 60\% (with both source and target having 20\% cropping), there is a noticeable reduction in registration accuracy. \ray{Futher directions include introducing $\SE(3)$-equivariant transformers into the encoder for capturing farther feature correlations.}
There are trade-offs with our method, including lengthy training times and reduced performance with limited overlap. Our unsupervised training approach uses a curriculum learning strategy that progresses from smaller to larger rotations, resulting in extended training times. Training on the ModelNet40 dataset with 8 NVIDIA V100 GPUs takes a week for $90^{\circ}$ registrations. Performance decreases in low-overlap scenarios, as indicated in Table~\ref{exp_modelnet_benchmark}, where 60\% overlap (20\% cropping) reduces registration accuracy. However, data augmentation, longer training cycles, and a denser curriculum can improve low-overlap performance, while a sparser curriculum can reduce training time. Besides the extended training time, our method has an inference time of 0.03 to 0.1 seconds per iteration on a single V100 GPU. Future directions include incorporating $\SE(3)$-equivariant transformers into the encoder to capture more extended feature correlations, enhancing robustness and efficiency.

In summary, this paper introduces a differentiable, iterative point cloud registration framework that leverages correspondence-free pose regression in RKHS. \equivcvo achieves fine-grained feature space registration and effectively handles noise, outliers, and limited labeled data. Results on ModelNet40 and ETH3D datasets show our method outperforming established methods particularly in noise resilience. This study lays a foundation for further research in unsupervised equivariant learning within 3D vision and opens its application to numerous fields, including but not limited to robotics and the medical domain.

% ---- Bibliography ----
%
% BibTeX users should specify bibliography style 'splncs04'.
% References will then be sorted and formatted in the correct style.
%
\bibliographystyle{splncs04}
\bibliography{ieee-abrv,strings-abrv,main}

\end{document}